\documentclass[12pt,a4paper]{article}
\usepackage[left=2.54cm,top=2.54cm,right=2.54cm,bottom=2.54cm,bindingoffset=0.5cm]{geometry}
\usepackage{fontspec}
\usepackage{xeCJK}
\usepackage{unicode-math}

\usepackage{graphicx} 
\usepackage{caption}
\usepackage{subcaption}
\usepackage{adjustbox}

\usepackage{booktabs}
\usepackage{float}

\usepackage{fancyhdr}

\usepackage{listings} 

\usepackage{bm}
\usepackage{amsmath}

\usepackage{xcolor} 
\usepackage{hyperref}
\usepackage{comment}

\usepackage[backend=biber,natbib=true, style=apa]{biblatex}
\title{Phonetic and semantic analyses of spoken corpora of Beijing and Taiwan Mandarin indicate that the neutral tone is a lexical tone}

\author{Yuxin Lu$^{1}$, 
Zhexuan Li$^{2}$, 
R.Harald Baayen$^{3}$ \\ \ \\ \
$^{1}$ Quantitative Linguistics, Eberhard Karls Universität Tübingen,  \\
Tübingen, Germany \\
Email: yuxin.lu@uni-tuebingen.de \\
$^{2}$ Quantitative Linguistics, Eberhard Karls Universität Tübingen,  \\
Tübingen, Germany \\
Email: zhexuan.li@student.uni-tuebingen.de \\
$^{3}$ Quantitative Linguistics, Eberhard Karls Universität Tübingen, \\
Tübingen, Germany \\
Email: harald.baayen@uni-tuebingen.de \\
}
\date{June, 2026}

\providecommand{\keywords}[1]
{
  \small	
  \textbf{\textit{Keywords:}} #1
}

\begin{document}

\maketitle

\thispagestyle{empty}   

\newpage
\thispagestyle{empty} 

\begin{abstract}
\noindent
The neutral, or floating, tone of Mandarin Chinese is a tone with an enigmatic set of properties. It has been described as a reduced tone, or as a tone that sometimes is lexically fixed, but that can also be toneless. In two-syllable words, it is found only on the second syllable, but single-syllable words can also have the neutral tone.  We present a corpus-based study of the phonetic realization of the neutral tone in spontaneous conversational speech corpora of Beijing Mandarin and Taiwan Mandarin. We show that the neutral tone has its own tonal target, just as the four lexical tones of Mandarin.  We also show that disyllabic words with a neutral tone have pitch contours that have a pitch component that depends on the tone on the first syllable, just as has been observed for two-syllable words with a lexical tone on the second syllable \citep{Chuang:Bell:Tseng:Baayen:2026}. Furthermore, words with a floating tone have word-specific pitch signatures, which have also been documented for single-syllable words \citep{jin2026new} as well as two-syllable words \citep{lu2026realization}. These word-specific pitch signatures are shown to be predictable to some extent from words' contextualized embeddings, as previously reported for lexical tones \citep{Chuang:Bell:Tseng:Baayen:2026,lu2026realization}. As there is also considerable variability in the realization of lexical tones, we propose that the neutral tone is, in fact, a lexical tone in both Taiwan Mandarin and Beijing Mandarin.  We document both similarities and differences in the realization of the floating tone in these two varieties and provide evidence, using contextualized embeddings, that some of the observed differences may arise from differences in the meanings of the words as used in the two corpora.
\end{abstract}


\newpage

\setcounter{page}{1}    
\pagestyle{fancy}
\fancyhf{}
\fancyfoot[C]{\small\thepage}

\section{Introduction} 


\noindent
The tone system of standard Mandarin Chinese comprises four ``lexical" tones: a high level tone (T1), a rising tone (T2), a low dipping tone (T3), and a falling tone (T4). In addition, there is a fifth tone, the neutral tone (T5).  The neutral tone is reported to approach a mid-low pitch target, with a pitch contour that is largely determined by the preceding lexical tone \citep{chao1968grammar}. The phonetic realization of the neutral tone varies across regions, not only with respect to the details of how pitch is modulated, but also with respect to duration and intensity \citep{li2020tone}. 

Recent studies \citep{Chuang:Bell:Tseng:Baayen:2026, jin2026new, lu2026form,lu2026realization} have documented for unscripted conversational Taiwan Mandarin that words' pitch contours have a word-specific component that are tied to their meanings, independently of words' canonical tones and a wide range of prosodic factors.  These studies used the generalized additive model \citep[GAM,][]{wood_generalized_2017} to decompose empirical pitch contours into additive pitch contour components, each of which is tied to a linguistic predictor (e.g., the preceding tone, position in the sentence, or the tone of the following word). GAMs reveal strong evidence for both components tied to tone patterns and for word-specific pitch contour components.

\citet{lu2026realization} investigated the pitch realization of disyllabic words with as a set exemplify all possible 20 tone patterns (the four tones T1--T4, for the first syllable, and five tones (T1--T5) for the second syllable). In their analysis, the neutral tone emerged as a tone in its own right, along with the 4 lexical tones (T1--T4).  Given that the floating tone is described as highly dependent on context, this result is surprising. The present study therefore examines in further detail the pitch realization of disyllabic words in spontaneous speech, in which the first syllable bears a lexical tone and the second syllable carries a neutral tone. We address the following questions.

\begin{enumerate}
    \item  \textbf{What is the nature of neutral tone?} Do words with a specific initial lexical tone and a final neutral tone have a distinct pitch contour component that differs from the pitch contour components of words with other initial lexical tones?  This is what \citet{lu2026realization} observed for all 20 tone patterns, four of which had a neutral tone on the second syllable. If this result is replicated for Taiwan Mandarin, and also replicates for another variety of Mandarin, namely, Beijing Mandarin, this would imply that the realization of the neutral tone depends on the preceding tone in its own way. However, dependence on the preceding tone is also observed for final lexical tones in two-syllable compounds \citep[cf. for laboratory speech][]{xu1997contextual}.  Therefore, if evidence for tone-pattern specific pitch components for tone-patterns with a final floating tone replicates, this would argue in favor of the floating tone being no different from a lexical tone. 
    \item  Considerable differences have been reported for the realization of the floating tone in Taiwan Mandarin and Beijing Mandarin. Beijing Mandarin, for instance, has been reported to have a higher prevalence of neutral tones {\color{black} \citep{chirkova2011beijing, qian1985lun}}.  \textbf{Are these differences visible as differently-shaped pitch components for the same tone patterns (e.g., T2-T5) in Taiwan conversational speech and Beijing conversational speech?}
    Furthermore, is there more variability in the pitch signatures for tone patterns with the neutral tone in Beijing than in Mandarin?  In addition, it is possible that there is more variability in the realization of di-syllabic words with a neutral tone on the second syllable than for di-syllabic words with a lexical tone. 
    \item \textbf{Do disyllabic words with a neutral tone exhibit word-specific pitch contour signatures not only in Taiwan Mandarin, but also in Beijing Mandarin?} 
    If so, to what extent are the pitch signatures for a specific word similar across the two dialects?  One would expect there to be words with very similar pitch signatures, but for others, their pitch signatures may well be dialect-specific.
%
%
    \item \citet{Chuang:Bell:Tseng:Baayen:2026} and \citet{lu2026realization} have shown that words' pitch signatures can be predicted from their meanings (operationalized with contextualized embeddings) with above-chance accuracy, indicating that there is some isomorphy between the phonetic space of pitch contours and the embedding space.  This raises the following question:  \textbf{Are contextualized embeddings for Beijing conversational speech predictive for words' tonal signatures as previously observed for Taiwan Mandarin?} Given the substantial variability and greater prevalence of the neutral tone in Beijing Mandarin, this is not a trivial question. A related question is whether differences in words' pitch signatures in Beijing and Taiwan can be traced to differences in words' meanings in these two dialects, as gauged with contextualized embeddings.
\end{enumerate}

%
%
%
%
%
%
%

The remainder of this study is structured as follows. Section~\ref{sec:neutral_tone}  provides further background on the neutral tone in Standard Mandarin, Beijing Mandarin, and  Taiwan Mandarin. 
Section~\ref{sec:materials} introduces the speech corpora for Beijing and Taiwan and the datasets that we created from these corpora.  
Section~\ref{sec:analyses} presents the statistical analyses of the pitch contours, using generalized additive mixed modeling.  
Section~\ref{sec:modeling} reports our investigation of the relation between word-specific pitch signatures and word-specific meaning, using contextualized embeddings.  
Section~\ref{sec:discussion} provides the general discussion.

\section{The neutral tone}\label{sec:neutral_tone}

\noindent
The neutral tone occurs in unstressed syllables and is widely attested in grammatical morphemes. These include particles and suffixes (e.g., 了\ \textit{-le}, 个\ \textit{-ge}, 子\ \textit{-zi}), structural particles (的\ \textit{-de}), aspect markers (着\ \textit{-zhe}), locative elements (上\ \textit{-shang}), directional complements (来\ \textit{-lai}), as well as reduplicated forms (e.g., 妈妈\ \textit{ma1ma5}, `mom'). 

The neutral tone, also referred to as the  ``floating tone", is traditionally described \citep{chao1968grammar} as a separate tone category next to the four lexical tones. It is often analyzed as resulting from the neutralization of tonal contrasts in unstressed positions \citep{yan2024word}. It is also commonly referred to as qīng shēng (轻声) in Chinese, which literally means the ``light" or ``soft tone". Syllables with the neutral-tone are typically produced with shorter duration, lower intensity, less articulatory effort, and weakened vowel contrasts \citep{cao1986putonghua, chen2006production, wu2023mandarin}.


The pitch contours of floating tones have been reported to depend in part on the preceding lexical tone, and tend to converge to a mid or low pitch target by the end of the carrying syllable \citep{chao1968grammar, shih1987phonetics, yip2006tone}. Although the precise realization of the floating tone varies across studies, the pattern in the literature suggests that when following Tone 1, Tone 2, or Tone 4, the neutral tone tends to be realized with a falling contour or with an overall lower pitch than the offset of the preceding tone. However, when following Tone 3, it is often realized with a rising contour or with a pitch higher than the preceding offset \citep{chao1932preliminary, chao1968grammar, Chen_2000,shih1987phonetics, lee2008prosodic,cao1992neutral, dreher1968instrumental, lin1980beijinghua, yip2006tone, luo1957pu}.  

The realization of the neutral tone in connected speech is further co-determined by a wide range of factors, such as intonational context (e.g., declarative versus interrogative), prosodic structure (e.g., disyllabic versus trisyllabic sequences), the number of neutral tones within a sequence, and speech style. While sentence-final lexical tones in interrogative utterances typically exhibit a rising contour, sentence-final neutral tones may be realized with a falling contour in questions \citep{liu2007neutral}.  The realization of the neutral tone realization varies across disyllabic and trisyllabic sequences, with greater reduction often observed in longer sequences or in later syllables.  Sequences containing multiple neutral tones may show cumulative reduction effects in both duration and pitch \citep{xu2024cross, wu2023mandarin}. In addition, \citet{wu2023mandarin} reported that speech style further conditions neutral tone realization, with more extreme reduction typically observed in casual speech compared to more controlled styles such as broadcast news. 


\citet{huang2023qing} distinguished between nine different types of neutral tones, such as `obligatory neutral tone' words, `habitually neutralized' forms, `optionally' neutralizated words, and `contrastive neutral tone' words. Contrastive neutral words are found in the second syllable of certain disyllabic words, such as 东西\ \textit{dongxi} and 大意\ \textit{dayi}), which can be realized with different tone patterns depending on their meaning: \textit{dong1xi1} `east-west', \textit{dong1xi5} `thing'; \textit{da4yi4} `general idea', \textit{da4yi5} `careless'.


\subsection{Standard and Beijing Mandarin}

Standard Mandarin is based on Beijing Mandarin, but the two are not identical. As the official standard language of China, Putonghua takes the pronunciation of Beijing Mandarin as its norm while drawing on northern dialects for its grammatical foundation. Although the neutral tone is a common feature across northern Mandarin varieties, it is particularly prominent in Beijing Mandarin \citep{chirkova2011beijing, qian1985lun}, where it may also function as a sociolinguistic index of local identity \citep{zhang2005chinese}.  In Standard Chinese, about 15–20\% of the syllables in written texts are considered unstressed, including certain suffixes, clitics, and particles. Beijing Mandarin, however, has a particularly high prevalence of neutral tones. 

According to \citet{dong2025neutral}, the neutral tone should be understood as a flexible tone on a continuum from fully specified lexical tones to toneless syllables, with its actual realization shaped by contextual factors. Within this continuum,  the so-called ``forbidden'' neutral tone words are of special interest. Words with the forbidden neutral tone are often neutralized in Beijing Mandarin, but are prescribed to carry full lexical tones in Standard Mandarin. 
%
%
As a consequence, the Beijing dialect contains a larger inventory of neutral-tone words than Standard Mandarin \citep{Hu:1987}. 
%
%
%

Phonetically, in Beijing Mandarin, syllables with neutral tones have been found to be more reduced than in many other Mandarin varieties, exhibiting shorter duration and lower intensity \citep{lu1995putonghua}. Similar to standard Mandarin, their F0 realization is highly dependent on the preceding lexical tone, both in contour shape and pitch height \citep{lee2003phonetic}.




The northern varieties of Mandarin are relatively homogeneous compared to the considerable diversity observed among southern varieties. While most regional varieties, such as Guangzhou, Shanghai, Yantai,
share the same basic tonal categories as Beijing Mandarin, the phonetic realization of these tones, in particular their pitch contours, varies substantially across regions \citep{li2020tone} and serves as an important marker of one's local dialect.
%

The phonetic realization of the neutral tone has been investigated for a range of Mandarin varieties, including Changde Mandarin \citep{zhang2020neutral}, Changsha Mandarin \citep{xu2025plastic}, Ürümqi Mandarin \citep{hsieh2008study, wei2011Wulumuqi}, Taiwan Mandarin \citep{huang2018phonological}, Yichang Mandarin \citep{li2018disyllabic}, and Tianjin Mandarin \citep{li2019prosodically}. \citet{li2016production} provides a comparative analysis of Chongqing, Kunming, and Nanjing Mandarin.  Differences in neutral tones between Changsha Mandarin and Standard Mandarin \citep{xu2024cross} demonstrate variation across social strata and speech levels.  Whereas second syllables of some disyllabic words are also unstressed in Northern Mandarin accents,  many Mandarin speakers in Southern China tend to preserve their inherent tones.

\subsection{Taiwan Mandarin}

Taiwan Mandarin, which has been influenced by Southern Min, differs from Standard Mandarin in several respects. One notable characteristic is that the distinction between stressed and unstressed syllables is less perceptually salient \citep{kubler1985influence}. \citet{li2005preliminary} reported that the neutral tone in Taiwan Mandarin exhibits a relatively stable pitch target,  \citet{tseng2004prosodic} described the neutral tone in Taiwan Mandarin as resembling a low ``entering tone" with short duration.  \citet{huang2018phonological} reported that in Taiwan Mandarin the neutral tone may either extend the pitch contour of the preceding tone or converge toward a relatively stable pitch target in the mid-low to low range.  As not all syllables lacking lexical stress are necessarily reduced in the same way, \citet{huang2018phonological} argues that the neutral tone in Taiwan Mandarin is better analyzed as unaccented rather than strictly unstressed.

%

The use of neutral tone is reported to be less frequent in Taiwan Mandarin than in Standard Mandarin and Beijing Mandarin. \citet{huang2018phonological} compared the Putonghua Shuiping Ceshi Qingsheng Cibiao (Standard Mandarin Proficiency Test wordlist) with the Revised Mandarin Chinese Dictionary published by Taiwan's Ministry of Education and observed that while suffixes and reduplicated syllables are consistently marked as having a neutral tone, only about half of the syllables that have neutral tones in Standard Mandarin carry a neutral tone in Taiwan Mandarin. For example, the experiential aspect suffix 過  and directional complements such as 上 `up' and 來 `come' are often pronounced with their canonical tones (\textit{guo4}, \textit{shang4}, and \textit{lai2}) rather than as reduced syllables.

%
%
%
%
%
%
%
%


As for words with reduplicated syllables, only kinship terms are produced with a neutral tone in Taiwan Mandarin. However, to express endearment, the neutral tones of kinship terms can be realized as rising or high tones, as in 妈妈 \textit{ma3ma2} (`mother') and 姐姐 \textit{jie3jie1} (`older sister').  For verbal and nominal reduplication, the tone of the initial syllable is often maintained: 看看 (`take a look at') is realized as \textit{kan4kan4} rather than as \textit{kan4kan5} \citep{hsu2006revisiting}. 



\subsection{The semantics of the floating tone}

\begin{figure}[htbp]
  \centering
  \includegraphics[width=\linewidth, height=7cm]{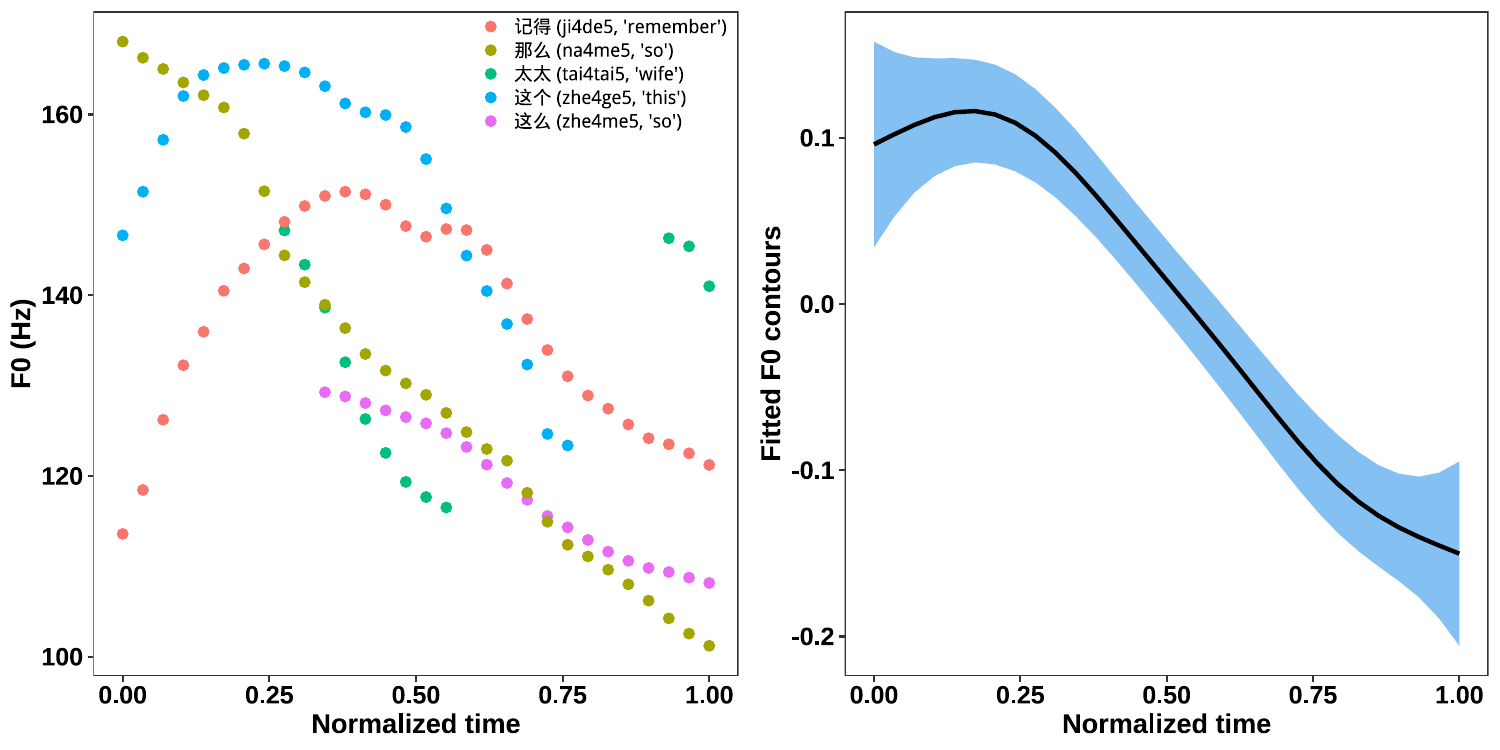}
  \caption{Toy dataset. The left panel shows the F0 contours of five selected tokens with T4-T5 tone pattern, produced by the same speaker, from the corpus of spoken Beijing Mandarin. The right panel shows the fitted F0 contours predicted by a simple GAM, using a thin plate regression spline smooth for normalized time as predictor.}
  \label{fig:toy}
\end{figure}

As mentioned above, several studies have reported that in conversational Taiwan Mandarin, words do not only have a pitch component that reflects the canonical tones, but also a pitch component that most likely reflects words' semantics  \citet{Chuang:Bell:Tseng:Baayen:2026,jin2026new,lu2026realization}. The statistical tool that led to the discovery of word-specific tonal signatures is the generalized additive model \citep{wood_generalized_2017}. Figure~\ref{fig:toy} illustrates the component reflecting words' tone pattern for a toy example with 5 words with the T4-T5 tone pattern.  The left panel presents words' individual pitch contours in normalized time. The right panel shows the pitch component that the GAM reconstructs as being shared by all five words.  Individual words deviate from this shared pitch component.  When pitch data are available for many word tokens of the same word type, then it becomes possible to not only isolate the pitch component shared by all tokens of all word types with a given tone pattern, but to also isolate for each word type what the characteristic tone component of that word type is.  

\begin{figure}[htbp]
    \centering
    \includegraphics[width=\linewidth]{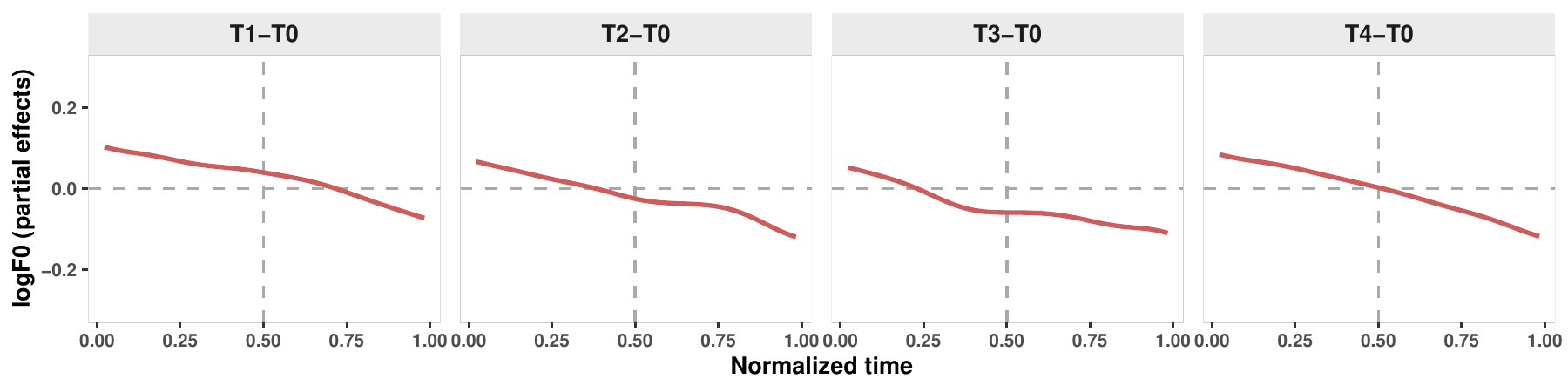}
    \caption{
    The average predicted pitch contours for the T1-T5, T2-T5, T3-T5, and T4-T5 tone patterns. The contours are reproduced from the black curves in Figure~5 of \citet{lu2026realization}.
    }
    \label{fig:lu2026}

    \vspace{0.5em}

    \begin{minipage}{0.95\linewidth}
    \footnotesize
    \textit{Note.} \citet{lu2026realization} labels the neutral tone as T0, whereas the present study uses T5. Accordingly, the original tone patterns T1-T0, T2-T0, T3-T0, and T4-T0 are presented here as T1-T5, T2-T5, T3-T5, and T4-T5.
    \end{minipage}
    
\end{figure}

It is of course essential to take into account the many different prosodic factors that have been shown to modulate the phonetic realization of pitch. Figure~\ref{fig:lu2026} presents the pitch components of the tone patterns with a final floating tone, isolated by a GAM for the dataset of conversational Taiwan Mandarin studied by \citet{lu2026realization}. These pitch components are modulated by the preceding and following tones (3\ldots 4, 4\ldots 0, 4\ldots 1, 4\ldots 4), represented by shades of blue. The average of these components is shown in red. The overall trend appears to be a downward sloping pitch contour that is rather similar irrespective of the lexical tone of the first syllable. The GAM of \citet{lu2026realization}, and the GAM models to be presented below, take into account many more prosodic factors (duration, speaker, gender, position in the utterance,  bigram probability), in order to optimize estimates of the pitch components tied to tone patterns and the pitch signatures tied to individual word types.

In the study of \citet{lu2026realization}, tokens and types of words with a neutral tone on the second syllable were less frequent than the types and tokens of words with tone patterns consisting of only lexical tones.  We therefore carried out a follow-up study for Taiwan Mandarin, which we complemented with a replication study for Beijing Mandarin.

\section{Materials}\label{sec:materials}

\noindent
The data analyzed in the current study are drawn from spontaneous speech corpora representing two varieties of Mandarin: Beijing Mandarin and Taiwan Mandarin. This section provides a detailed description of these  corpora, acoustic processing, and data sampling. 

\subsection{Corpora}

\noindent
The data for Beijing Mandarin data were drawn from a corpus of conversational speech \citep{Mandarincorpus}, consisting of recordings from 50 speakers (25 female, 25 male).  
All recordings were conducted in a soundproof recording studio in Beijing using high-quality audio equipment.

The Taiwan Mandarin data were extracted from the Taiwan Mandarin Spontaneous Speech Corpus \citep{fon_preliminary_2004}, which contains approximately 30 hours of free and unstructured interview speech. This corpus includes 55 native speakers of Taiwan Mandarin (31 female, 24 male) aged between 20 and 60 years. Recordings were collected in Taipei over several years between 2000 and 2010 using high-quality microphones and digital recorders in quiet locations, preferably in soundproof laboratories. Before the interviews, participants were informed that the study aimed to investigate their views on various aspects of life. Interviewers minimized active participation in order to elicit extended monologues, covering topics such as childhood, education, work, personal relationships, and other domains of life. After the interview, participants were asked for consent for the use of their recordings in research. 

Both corpora followed comparable elicitation strategies designed to encourage long stretches of spontaneous speech while minimizing interviewer intervention. As a result, the recordings consist of naturally occurring speech covering broadly comparable thematic domains. 

To further assess comparability in discourse content, we made use of topic modeling, an unsupervised machine learning technique for identifying latent thematic structure. We used the BERTopic framework \citep{grootendorst2022bertopic} to model topics in the corpora. Prior to modeling, the data were preprocessed through tokenization and the removal of stopwords, using corpus-specific stopword lists for Beijing Mandarin (Simplified Chinese) and Taiwan Mandarin (Traditional Chinese). 
Topic interpretation was based on the highest-weight (i.e., most representative) words within each topic. Results indicate that both corpora cover a wide range of social, cultural, technological, and personal-experience-related topics, although the specific thematic emphases may differ. In the Beijing Mandarin corpus, prominent topics include national and regional topics, mobile technology and digital consumption, travel and sports, urban public transportation infrastructure, everyday mobility, and emerging technologies. The Taiwan Mandarin corpus features frequently discussed such as health and medicine, marriage and family, music, education and extracurricular activities, natural disasters and environmental risk, and leisure activities such as golf. 

\subsection{Transcription and selection criteria}

\noindent
The audio recordings of Beijing Mandarin were transcribed and annotated using simplified Chinese characters. Forced alignment between the audio and transcriptions was carried out using the Montreal Forced Aligner \citep[MFA,][]{mcauliffe2017montreal}. Pauses, which commonly occur in spontaneous speech, were identified and labeled as empty intervals. The resulting automatic segmentation was subsequently manually checked by native speakers. Fewer than 5\% of tokens showed incorrect or imprecise alignments and were subsequently corrected, following established practices in previous studies \citep{sun2021boundary}.

The Taiwan Mandarin recordings were transcribed using traditional Chinese characters. Word boundaries in the orthographic representation were identified using a word segmentation system \citep{ma2003introduction}, and canonical tone information was assigned based on dictionaries. The character transcriptions were then romanized to enable forced alignment with Easyalign \citep{goldman2011easyalign} on character and word levels. Transcriptions were aligned at the word and syllable levels with the audio. The resulting alignment was manually checked and, where necessary, corrected.

Since Beijing and Taiwan Mandarin differ with respect to which words carry floating tones,\footnote{
E.g., 厉害\ (`severe; impressive') is listed as \textit{li4hai5} (T4–T5) in Beijing standard, but as \textit{li4hai4} (T4–T4) in Taiwan standard
} for our cross-dialect comparison, we decided to select words based on the tones as indicated in \textit{Xiandai Hanyu Cidian} (\textit{现代汉语词典}) [the Contemporary Chinese Dictionary (7th edition)] \citep[][]{CASS2016}, as this maximizes the opportunity to find words that are, at least in one of the dialects, realized with a floating tone.  We also extracted the normative tones for Taiwan Mandarin, using the Ministry of Education Dictionary (\textit{教育部国语辞典简编本}), a widely used and regularly updated reference for contemporary, frequently used vocabulary in Taiwan Mandarin.
%
%
Our analysis focused on disyllabic words with a lexical tone (T1, T2, T3 and T4) on the first syllable and a neutral tone (Tone 5) on the second syllable, which corresponds to the T1-T5, T2-T5, T3-T5, T4-T5 tone patterns. All selected words in the Beijing dataset conform to this pattern according to the Beijing standard. However, some of these words are listed with a full lexical tone on the second syllable in Taiwan Mandarin dictionaries. Importantly, to preserve cross-variety comparability, such items were retained in the Taiwan dataset but grouped into a separate category labeled ``Others''. This category comprises those words that are classified as neutral-tone items in Beijing Mandarin but as bearing a full lexical tone in Taiwan Mandarin.

For each of the two corpora, we selected all tokens of two-syllable word types with a neutral tone (according to the authoritative dictionaries of their respective dialects). Following \citep[e.g.,][]{Chuang:Bell:Tseng:Baayen:2026,lu2026form, lu2026realization}, we selected only those word types for which at least five tokens were available in the corpus, to ensure sufficient representation for statistical modeling. Furthermore, in order to prevent model prediction being dominated by a few highly frequent words, the maximum number of tokens per word type was capped at 200. In addition, we imposed the requirement that each word type was produced by at least one female and one male speaker to avoid gender- and speaker-based imbalance.

As shown in Table~\ref{tab:dataset}, the resulting Beijing Mandarin dataset contains 4,871 tokens representing 101 word types across four tone patterns (T1-T5, T2-T5, T3-T5, and T4-T5). The Taiwan Mandarin dataset contains 3,831 tokens representing 80 word types, across five categories (T1-T5, T2-T5, T3-T5, T4-T5, and \texttt{Others}). In general, the mean number of tokens per type is comparable between the two datasets. 

\begin{table}[htbp]
    \centering
    \caption{Summary of tokens and word types in each tone pattern, including the shared subset for Beijing and Taiwan Mandarin.
    }
    \begin{tabular}{lcccc}
        \toprule
        \textbf{Corpus / Tone pattern} & \textbf{Tokens} & \textbf{Word types} & \textbf{Mean number of tokens per type} \\
        \midrule
        \textbf{Beijing} & & & \\
        T1-T5 & 854 & 26 & 32.85 \\
        T2-T5 & 1558 & 24 & 64.92 \\
        T3-T5 & 1010 & 18 & 56.11 \\
        T4-T5 & 1449 & 33 & 43.91 \\
        \textbf{Total} & \textbf{4871} & \textbf{101} & \textbf{48.23} \\
        \addlinespace[0.15cm]
        
        \textbf{Taiwan} & & & \\
        T1-T5 & 727 & 10 & 72.70 \\
        T2-T5 & 565 & 13 & 43.46 \\
        T3-T5 & 636 & 11 & 57.82 \\
        T4-T5 & 866 & 20 & 43.30 \\
        Others & 1037 & 26 & 39.88 \\
        \textbf{Total} & \textbf{3831} & \textbf{80} & \textbf{47.89} \\
        \bottomrule
    \end{tabular}
    \label{tab:dataset}
\end{table}

The ``Others" category contains 26 word types (see Table~\ref{tab:other}). Although many of these words are realized as T5 in Beijing Mandarin, they are classified as Others here because the Taiwan Mandarin Dictionary specifies a full lexical tone rather than Tone~5 on the second syllable.

\begin{table}[htbp]
\centering
\caption{The 26 word types in the \textit{Others} category of the Taiwan Mandarin dataset. These are disyllabic words whose second syllable does not carry T5. Pinyin follows the Taiwan Mandarin Dictionary.}
\begin{tabular}{lll}
\toprule
Word & Pinyin (Taiwan Mandarin Dict) & English translation \\
\midrule
便宜 & \textit{pian2yi2} & inexpensive; cheap \\
部分 & \textit{bu4fen4} & part; portion \\
部份 & \textit{bu4fen4} & part; portion \\
告訴 & \textit{gao4su4} & tell; inform \\
故事 & \textit{gu4shi4} & story \\
護士 & \textit{hu4shi4} & nurse \\
厲害 & \textit{li4hai4} & severe; impressive \\
麻煩 & \textit{ma2fan2} & trouble; bother \\
模糊 & \textit{mo2hu2} & blurry; vague \\
朋友 & \textit{peng2you3} & friend \\
漂亮 & \textit{piao1liang4} & beautiful; pretty \\
親戚 & \textit{qin1qi1} & relative \\
清楚 & \textit{qing1chu3} & clear \\
情形 & \textit{qing2xing2} & situation; condition \\
熱鬧 & \textit{re4nao4} & lively; bustling \\
認識 & \textit{ren4shi4} & know; recognize \\
時候 & \textit{shi2hou4} & time; moment \\
事情 & \textit{shi4qing2} & matter; affair \\
舒服 & \textit{shu1fu2} & comfortable \\
喜歡 & \textit{xi3huan1} & like \\
先生 & \textit{xian1sheng1} & Mr.; gentleman \\
想想 & \textit{xiang3xiang3} & think about it \\
消息 & \textit{xiao1xi2} & news; information \\
休息 & \textit{xiu1xi2} & rest \\
知識 & \textit{zhi1shi4} & knowledge \\
做作 & \textit{zuo4zuo4} & affected; artificial \\
\bottomrule
\end{tabular}
\label{tab:other}
\end{table}

\subsection{Acoustic processing}

%

\noindent
F0 extraction was carried out using Praat \citep{boersma_praat_2020}, implemented via the \textit{Parselmouth} Python interface. For each token, F0 was measured across the entire syllable, including onset consonants, vowels, and final nasals, if present. The pitch floor and ceiling were set to 75–400 Hz for female speakers and 50–300 Hz for male speakers. No smoothing or interpolation was applied in order to preserve the original acoustic speech signal. F0 values were not computed in cases where voicing was absent (e.g., voiceless consonants) or where creaky voice prevented reliable estimation; these values were coded as missing (\textit{NA}). Abrupt pitch jumps between consecutive time points, probably due to F0 tracking errors, were excluded. Tokens with potential F0 tracking errors or creaky voice were identified through visual inspection by native speakers and excluded from further analysis.  Thus, despite differences in the alignment tools used for the two corpora, extensive manual inspection of the tokens selected for analysis with respect to alignment accuracy ensured comparability across the two datasets.

\subsection{Predictors}

For statistical analysis, we collected the following predictors:
\begin{description}
    \item [normalized time] For each word token, time was normalized between 0 and 1, allowing tokens with different durations to be modeled on a common time scale.
    \item [tone pattern] The tonal pattern of two-syllable word, the combination of the lexical tone of the first syllable and the neutral tone of the second syllable. In the Beijing Mandarin dataset, this factor has four levels: \textit{T1-T5}, \textit{T2-T5}, \textit{T3-T5}, and \textit{T4-T5}. In the Taiwan Mandarin dataset, it has five levels:  \textit{T1-T5}, \textit{T2-T5}, \textit{T3-T5}, and \textit{T4-T5}, and \textit{Others}. 
    \item [word] The orthograghy of word, which is a factor with 101 levels in the Beijing dataset and a factor with 80 levels in the Taiwan Mandarin dataset.
    \item [speaker] A unique speaker identifier, a factor with 50 levels for the Beijing dataset and with 55 levels for the Taiwan Mandarin dataset.
    \item [duration] A continuous variable indicating the duration of a token, measured in seconds.
    \item [word position in utterance] The normalized position of a word within its utterance. This is calculated by dividing the word's position by the total number of syllables in the utterance, yielding a value between 0 and 1. Lower values indicate that the token occurs closer to the beginning of the utterance, and higher values indicate that it occurs closer to the end of the utterance. For single-word utterances, the position is coded as 1. Utterance is defined as a continuous stretch of speech bounded by silent intervals in the corpus segmentation.
    \item [tonal context] A factor with 36 levels, comprising all combinations of the tone on the syllable preceding the word and the tone on the syllable following the word. If a pause occurs immediately before or after a word, the corresponding tone is coded as \texttt{PAUSE}. Speakers occasionally engage in code-switching, typically between English and Chinese. For tokens immediately preceded or followed by an English word, no tone could be reliably identified; these tokens were therefore labeled as \texttt{NA} and excluded from further analyses. With five tones (T1, T2, T3, T4, and T5) and the \texttt{PAUSE}, and two positions (preceding and following), the number of levels of tonal context is 6$\times$ 6 = 36.
    \item [bigram probability of the previous word] Bigram probability quantifies how predictable a word is in its context. This measure of contextual predictability is based on the relative frequency of the word's co-occurrence with surrounding words. A higher bigram probability indicates that the target word is more predictable within its given context. In the present study, \texttt{bg\_prob\_prev} is calculated as the probability of the occurrence of the target word given the preceding word.
    \item [bigram probability of the following word] This measure represents the bigram probability of the following word, calculated as the probability of the occurrence of the target word given the following word.
\end{description}

\section{Analysis}\label{sec:analyses}

\noindent
We made use of the Generalized Additive Model \citep[GAM,][]{wood_generalized_2017,chuang2021analyzing} to decompose pitch contours over normalized time into a series of component contours, using the implementation of the \texttt{mgcv} package \citep{wood_generalized_2017} for R \citep{team_r_2020}. We implemented an AR(1) process (first-order autoregressive model) in the residuals to take into account that there are strong autocorrelation in time series of pitch measurements.  The inclusion of the AR(1) process with an autocorrelation coefficient of $\rho = 0.90$ substantially reduced residual autocorrelation. 


The response variable was the natural logarithm-transformed F0.\footnote{
The log transformation resulted in a distribution of F0 values that roughly followed a normal distribution.  We did not use MEL or BARK scales, as our interest is primarily in the production of pitch contours, rather than the comprehension of pitch contours.
} Pitch contours were modeled over entire words, including onset consonants, vowels, and final nasals, if present.\footnote{For detail discussion of GAMs reconstruct pitch contours exactly as desired for voiceless segments, see \citet{Chuang:Bell:Tseng:Baayen:2026}.} 


We analyzed the Beijing and Taiwan datasets separately. For each variety, we started with a simple baseline model that included a smooth for normalized time, as well as by-speaker random nonlinear components, using a factor smooth with shrinkage. These by-speaker components account for differences in how speakers, on average, realize pitch contours. 

The baseline GAM was specified as follows:

\begin{equation}
\begin{aligned}
\texttt{logpitch} \sim\;&
    \texttt{s(normalized\_t, k = 5)} \\
    &+ \texttt{s(normalized\_t, speaker, bs = 'fs', m = 1)} \\
\end{aligned}
\end{equation}
We then fitted 7 additional GAMs, each with one additional predictor (\texttt{word}, \texttt{tonal context}, \texttt{tone pattern}, \texttt{word position}, \texttt{duration}, and \texttt{bigram probability previous} and \texttt{bigram probability following}). For \texttt{word}, \texttt{tonal context} and \texttt{tone pattern} we implemented factor smooth interactions with normalized time. For the covariates, we implemented tensor product smooths to model their interactions with normalized time. As many of these predictors are correlated, these simple models make it straightforward to assess the relative variable importance of the individual predictors. We assessed variable importance by calculating the extent to which model fit improved using Akaike's Information Criterion (AIC): greater reductions in AIC imply improved model fit.  

\begin{figure}[htbp]
    \centering
    \includegraphics[width=\linewidth]{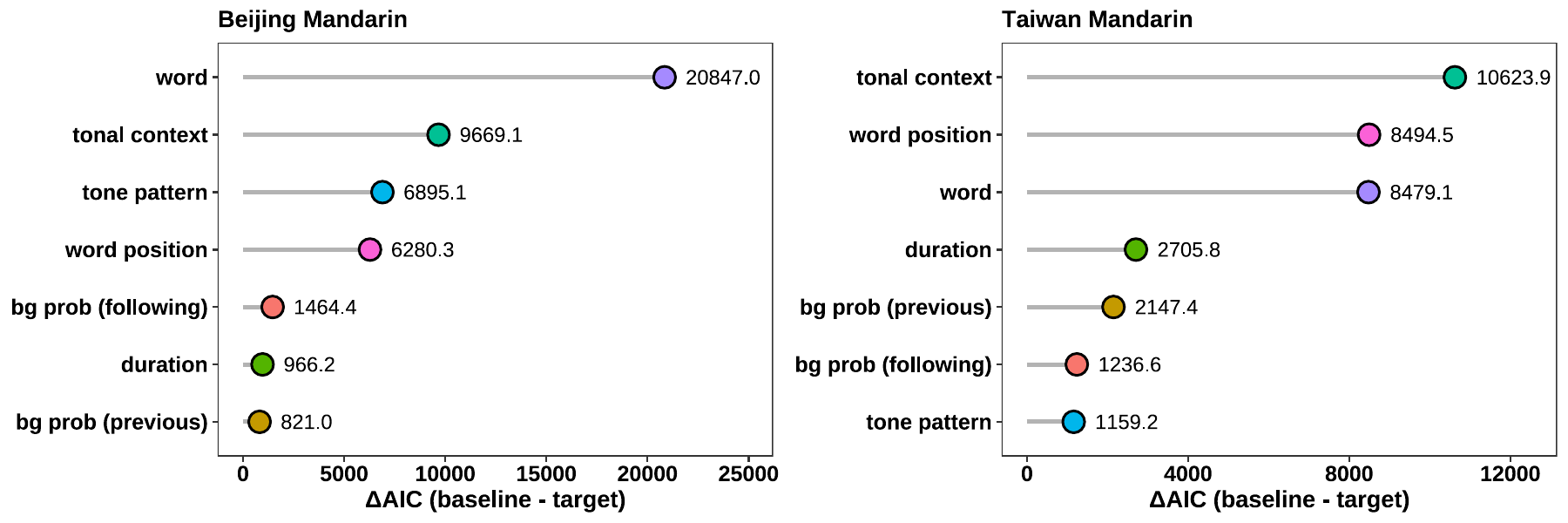}
    \caption{Reduction in AIC when a predictor of interest is added to the baseline GAM. Larger reductions indicate greater variable importance.  Results are shown separately for Beijing Mandarin (left panel) and Taiwan Mandarin (right panel).}
    \label{fig:aic}
\end{figure}

Figure~\ref{fig:aic} presents the reductions in AIC obtained for each of the predictors, for Beijing Mandarin (left panel) and Taiwan Mandarin (right panel).  In Beijing Mandarin, \texttt{word} yielded a substantially greater improvement in model fit than any other predictor, with an AIC reduction of 20874 units. \texttt{Tonal context} ranked second, followed by \texttt{tone pattern} and \texttt{word position}.  In contrast, for Taiwan Mandarin, \texttt{tonal context} emerged as the most important predictor, followed by \texttt{word position} and \texttt{word}. Comparing the results for the two dialects, \texttt{word} is the more important predictor for the Beijing dataset, followed by \texttt{tonal context}, whereas for the Taiwan dataset, \texttt{tonal context} is the most important variable, followed by \texttt{word} and \texttt{word position}. 


We then fit GAMs including all predictors, using the following model specification:

\begin{equation}
\begin{aligned}
\texttt{logpitch} \sim\;&  \texttt{tone\_pattern} \\
    &+ \texttt{s(normalized\_t, by = tone\_pattern, k = 5)} \\
    &+ \texttt{s(normalized\_t, speaker, bs = 'fs', m = 1)} \\
    &+ \texttt{s(normalized\_t, word, bs = 'fs', m = 1)} \\
    &+ \texttt{s(normalized\_t, tonal\_context, bs = 'fs', m = 1)} \\
    &+ \texttt{s(norm\_utt\_pos, k = 5)} \\
    &+ \texttt{ti(normalized\_t, norm\_utt\_pos, k = c(4,4))} \\
    &+ \texttt{s(logdur, k = 8)} \\
    &+ \texttt{ti(normalized\_t, logdur, k = c(5,5))} \\
    &+ \texttt{s(bg\_prob\_prev, k = 5)} \\
    &+ \texttt{ti(normalized\_t, bg\_prob\_prev, k = c(5,5))} \\
    &+ \texttt{s(bg\_prob\_fol, k = 5)} \\
    &+ \texttt{ti(normalized\_t, bg\_prob\_fol, k = c(5,5))} \\
\end{aligned}
\end{equation}

\begin{figure}[htbp]
    \centering
    \includegraphics[width=\linewidth]{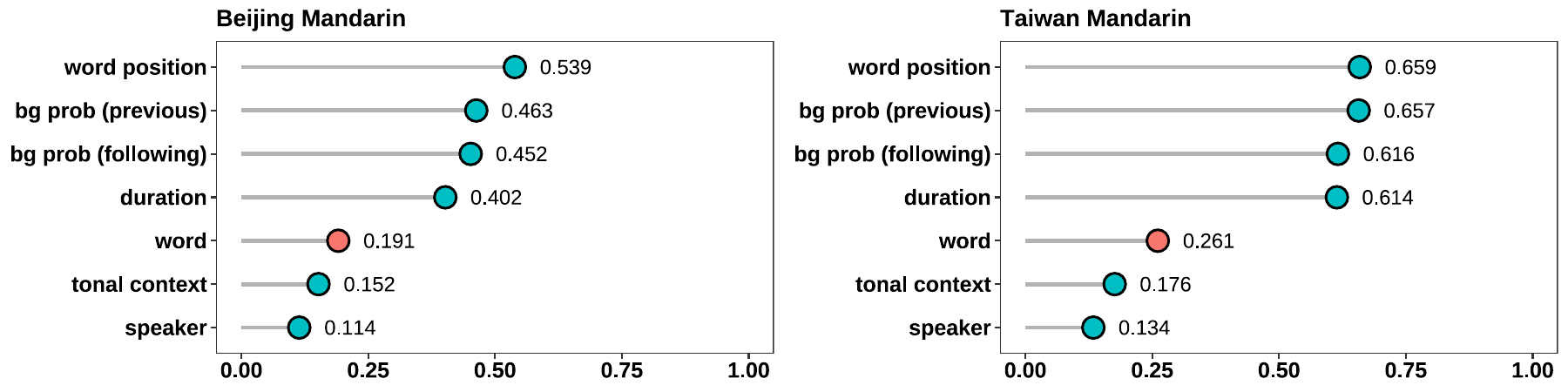}
    \caption{
    Estimated concurvity scores for smooth terms in the best-fit GAMs for Beijing Mandarin (left panel) and Taiwan Mandarin (right panel). Lower concurvity values indicate that a predictor is contributing more independently to the model fit. The concurvity scores for the \texttt{word} smooth are highlighted in red.
    }
    \label{fig:concurvity}

    \vspace{0.5em}

    \begin{minipage}{0.95\linewidth}
    \footnotesize
    \textit{Note.} For models with continuous predictors, only the concurvities of smooth terms involving normalized time are shown. The relatively high concurvity values observed for tone-pattern smooths are unsurprising given the substantial similarity among the pitch trajectories associated with the four tone patterns (cf. Figure~\ref{fig:pattern}).
    \end{minipage}
\end{figure}

This model improved substantially on the single predictor GAMs (decrease in AIC compared to the best fitting single predictor models: 32704.17 units for the Beijing dataset and 28593.11 units for the Taiwan dataset).
This complex model, which brings a wide range of prosodic factors under statistical control, has one downside. Due to the correlations between predictors, there is some concurvity in the model. High concurvity scores indicate that the partial effect of one predictor is predictable from other predictors in the model, rendering model interpretation more difficult. Figure~\ref{fig:concurvity} presents the estimated concurvity scores for Beijing Mandarin (left panel) and Taiwan Mandarin (right panel).
The concurvity scores are highest for four control variables for prosody: \texttt{word position, duration, previous bigram probability}, and \texttt{following bigram probability}.  Unsurprisingly, concurvity scores are the lowest for \texttt{speaker}, followed by \texttt{tonal context}.  Concurvity scores are also relatively low for \texttt{word}.  The low concurvity scores for \texttt{word} and \texttt{tonal context} indicate that the partial effects of these predictors are interpretable, as part of a larger model that controls for  prosodic factors. 


\begin{figure}[htbp]
    \centering
    \includegraphics[width=\linewidth]{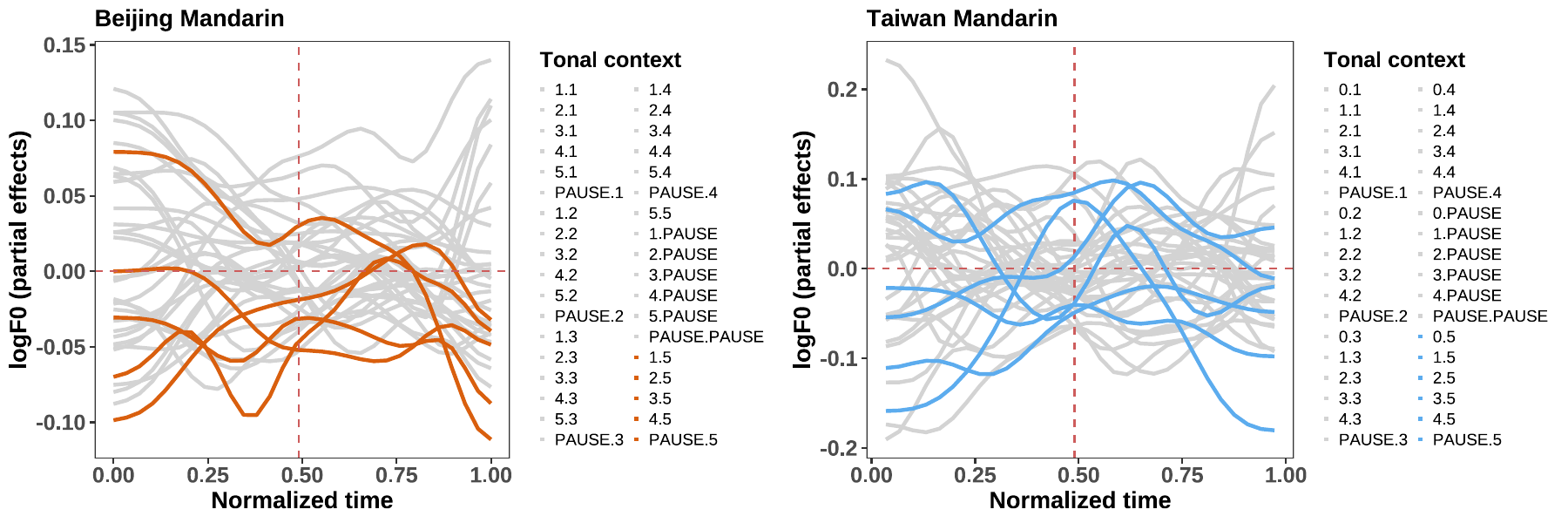}
    \caption{
    The modulation of tonal context on predicted pitch contours, showing all 36 tonal context levels in Beijing Mandarin (left panel) and Taiwan Mandarin (right panel). The predicted pitch contours are estimated from the partial effect of the factor smooth for tonal context. In both panels, contexts in which the neutral-tone syllable is followed by another neutral tone are highlighted by color. They represent tonal sequences T1–X–T5–T5, T2–X–T5–T5, T3–X–T5–T5, T4–X–T5–T5, and PAUSE–X–T5–T5, in which ``X–T5" represents an average of disyllabic word across four tone patterns. 
    All other tonal contexts are shown in grey.
    }
    \label{fig:context}
\end{figure}

Tonal context is a strong co-determinant of pitch contours in Taiwan Mandarin. Figure~\ref{fig:context} illustrates its effect on the predicted contours while holding the tone pattern constant. The left and right panels show the 36 levels of tonal context for Beijing Mandarin and Taiwan Mandarin, respectively. Interestingly, when a neutral tone is followed by another neutral tone (highlighted in red and blue, respectively), the predicted F0 at the end of the syllable is lower than in other tonal contexts, especially for Beijing Mandarin.

\noindent
In what follows, we proceed with examining the partial effects of tone pattern, and then consider the partial effect of word. 

\subsection{Tone pattern}

\begin{figure}[htbp]
    \centering    
    \includegraphics[width=\linewidth]{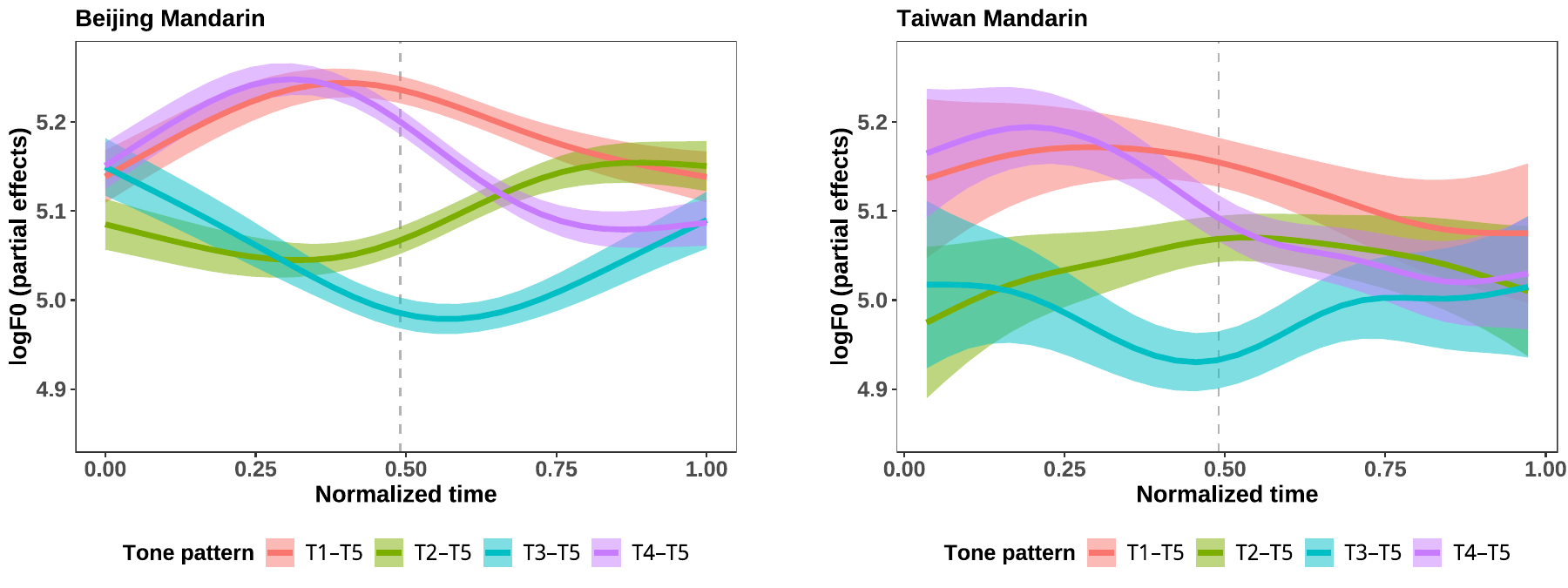}
    \caption{Predicted pitch contours for tone patterns in Beijing (left panel) and Taiwan (right panel).
    The pitch contours shown represent the partial effect of tone pattern, in combination with the corresponding tone-pattern specific intercepts. Separate GAMs were fitted to the two dialects.  
    The vertical dashed line indicates the average syllable boundary on the normalized time scale, with the first half corresponding to the initial lexical tone, and the second half corresponding to the neutral-tone syllable. They are obtained from two separate models for Beijing Mandarin and Taiwan Mandarin.
    }
    \label{fig:pattern}
\end{figure}

\noindent
Figure~\ref{fig:pattern} represents the predicted pitch contours for tone patterns (T1-T5, T2-T5, T3-T5, T4-T5) in Beijing (left panel) and Taiwan Mandarin (right panel). The first half of each contour before the average syllable boundary, corresponding to the normalized time interval [0, 0.49], represents the F0 contour of the lexical tone, while the second half [0.49, 1] represents that of the neutral tone syllable. The predicted pitch contours show narrower confidence intervals and higher pitch height in Beijing Mandarin than in Taiwan Mandarin. The narrower confidence intervals for Beijing Mandarin  dovetail well with the observation that tone pattern ranks as the third most important predictor in Beijing Mandarin, but as the least important in Taiwan Mandarin in the variable importance analysis (see Figure~\ref{fig:aic}). 

In Taiwan Mandarin, the first halves of the F0 contours largely resemble the prototypical citation forms of the respective lexical tones. T1 is high and level throughout the syllable, remaining higher than T2 and T3. T2 might be realized with a rising contour, but as the  confidence interval is very wide, the T2 of the Taiwan T2-T5 pattern might just as well be described as a mid level tone. T3 displays a falling trend, with the rise beginning around the average syllable boundary. T4 closely resembles its citation form with a clear fall. Overall, lexical tones in the stressed position of two-syllables largely maintain their canonical shapes.

In Beijing Mandarin, the realization of lexical tones shows greater deviation from citation forms. T1 and T4 have an initially rising F0 in Beijing Mandarin.  T2 exhibits a  small initial fall followed by a modest rise beginning around 70\% into the rhyme of the first syllable, and that continues in the second syllable.  T3 is realized with strong initial fall that continues slightly  into the second syllable.

\begin{figure}[htbp]
    \centering
    \begin{subfigure}[t]{0.49\linewidth}
        \centering
        \includegraphics[width=\linewidth]{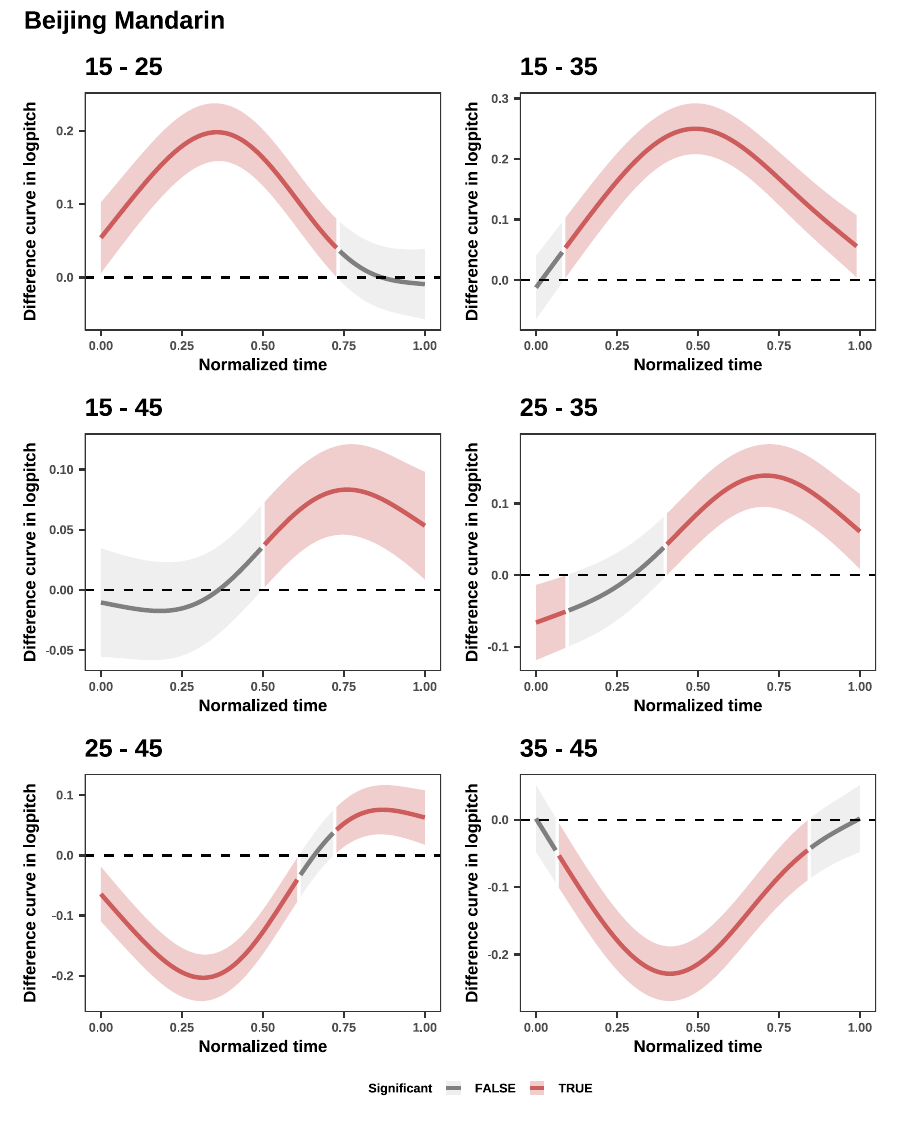}
        \caption{}
    \end{subfigure}
    \hfill
    \begin{subfigure}[t]{0.49\linewidth}
        \centering
        \includegraphics[width=\linewidth]{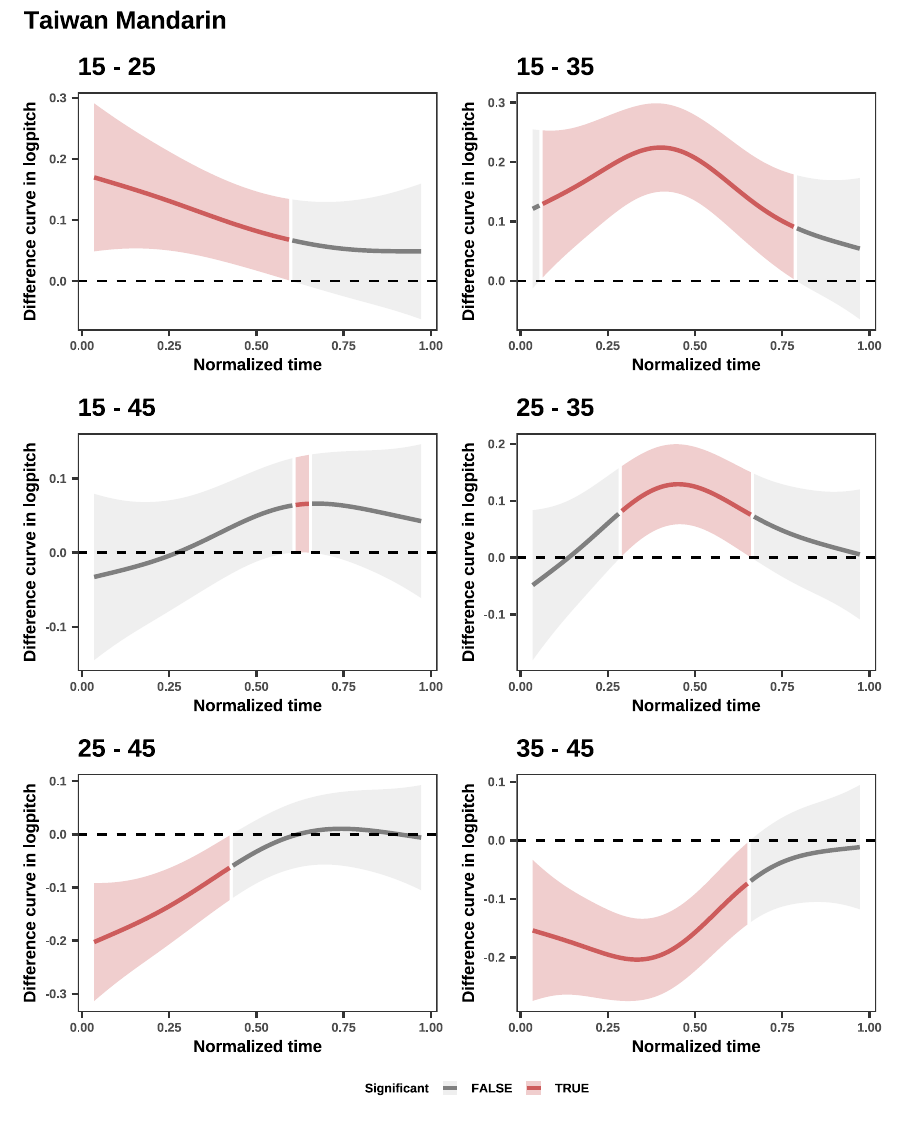}
        \caption{}
    \end{subfigure}
    \caption{Difference curves for pairs of tone patterns in Beijing Mandarin (left panel) and Taiwan Mandarin (right panel). The red area indicates that the normalized time domain where the estimated difference curve is significant different from zero, whereas grey area marks the normalized time domain with no significant difference.
    }
    \label{fig:diff}
\end{figure}

Figure~\ref{fig:diff} shows the estimated difference curve among pairs of tone patterns in Beijing Mandarin (left) and Taiwan Mandarin (right). These difference curves were  estimated by the \texttt{plot\_diff} function from \texttt{itsadug} package in R. The red areas indicate the normalized time intervals where the estimated difference curves are significantly different from zero, whereas the grey areas mark the time intervals where no significant difference is observed. The difference curve for T1-T5 and T2-T5 in Beijing Mandarin, as well as the difference curve for T3–T5 and T4–T5, includes zero at the endpoints of the pitch contours, indicating no significant differences in F0 at word offsets. All other pairs show significant differences in F0 at word offset. In Taiwan Mandarin, the confidence intervals of the estimated F0 differences approach zero toward the end of the contour across all comparisons. Consistent with earlier observations \citep{huang2018phonological}, this indicates that in Taiwan Mandarin, the neutral tone has a final mid-to-low target F0. 

The predicted contours of the neutral tone display systematic variation across tonal contexts in both Beijing and Taiwan Mandarin.  In both varieties, the F0 of the neutral tone generally approaches a final medium-to-low pitch target in both Beijing and Taiwan Mandarin. This final target appears to be somewhat lower for Taiwan Mandarin, but the wide confidence intervals argue for caution.  Unlike for Beijing Mandarin, all four tone patterns converge to the same word-final target, as shown by the difference curve analysis (see Figure~\ref{fig:diff}).

In the light of these findings, we can answer the first two research questions as follows.  With respect to \textbf{the nature of the neutral tone}, it is clear that lexical tones followed by a neutral tone have their own specific tone patterns that cannot be reduced to some general tone sandhi rule.   Some tone patterns can be interpreted as drawn-out versions of the initial lexical tone.  For Beijing Mandarin, the T2-T5 tone pattern could be interpreted as a rising tone with a late start of the rise. Likewise, the T3–T5 sequence in this variety could be seen as a  dipping tone that is realized over both syllables.  For Taiwan, the T3-T5 sequence can likewise be described as a dipping tone drawn out over both syllables.  For Taiwan Mandarin, furthermore, the T4-T5 pattern can also be seen as a falling tone, drawn out over both syllables, with a late start of the fall.  But other tone patterns have their own specific realizations, such as T1-T5 and T4-T5 in Beijing Mandarin, and T2-T5 in Taiwan Mandarin.  The only property shared by all T2 words is the mid-to-low final target. We take this property to be the defining characteristic of the T2-words, just as a final fall is a characteristic of two-syllable words with a T4 on the second syllable \citep[cf.][]{lu2026realization}.   

With respect to the question of \textbf{whether there are differences in the realization of tone patterns between Beijing and Taiwan Mandarin}, it is clear that marked differences indeed exist. The initial rise found for T1-T5 and T4-T5 words sets Beijing Mandarin apart from Taiwan Mandarin. The nearly level realization of T2-T5 sets Taiwan Mandarin apart from Beijing Mandarin. The tone pattern that is most similar across the two varieties is T3-T5.  Furthermore, there is more variability in the realization of tone patterns in Taiwan Mandarin, and in this variety, the word final tone target is the same for all four tone patterns. 

%

\subsection{Word}

\noindent
We next consider the question of whether words with a T2 on the second syllable have their own specific pitch signatures. The GAM analyses indeed provide strong support for word-specific tonal signatures not only for Taiwan Mandarin, but also for Beijing Mandarin. In fact, the variable importance for word in Beijing Mandarin is considerably stronger than that for Taiwan Mandarin (see Figure~\ref{fig:aic} above). 

\begin{figure}[htbp]
    \centering
    \begin{subfigure}[b]{\linewidth}
        \includegraphics[width=\linewidth]{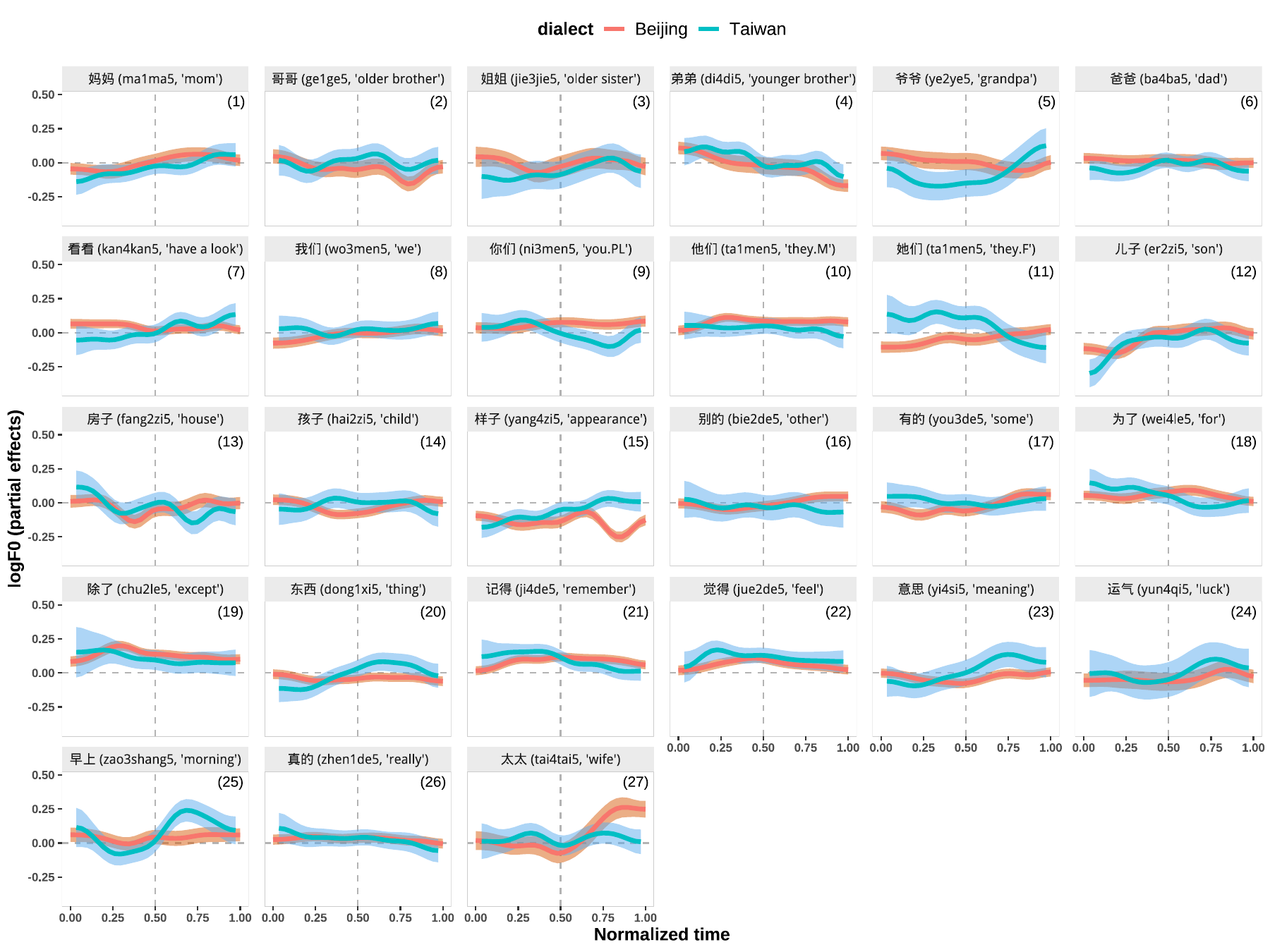}
        \caption{}
    \end{subfigure}
    \hfill
    \begin{subfigure}[b]{\linewidth}
        \includegraphics[width=\linewidth]{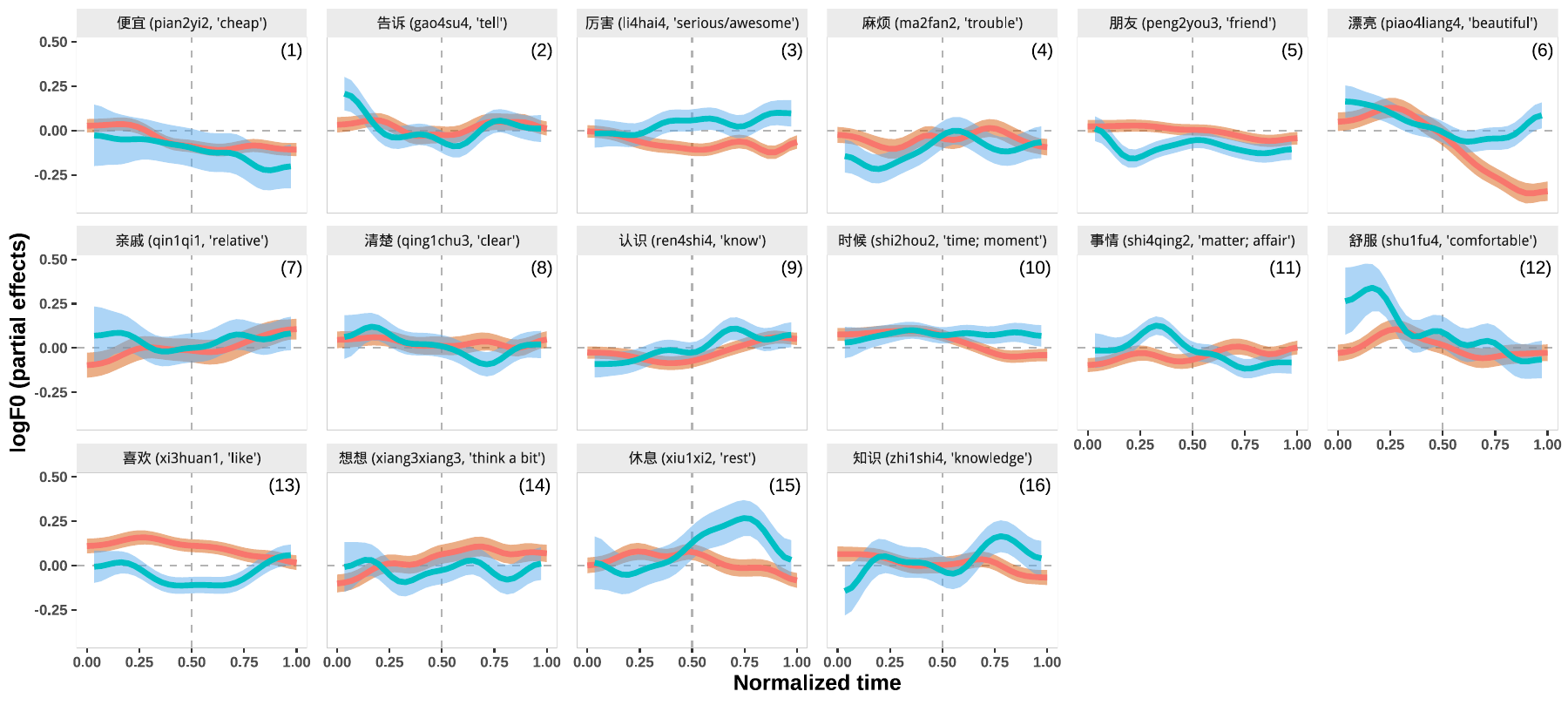}
        \caption{}
    \end{subfigure}
    \caption{Predicted pitch contours for individual words, estimated by combining the partial effect of the factor smooth for \texttt{word}, the partial effect of \texttt{tone pattern}, and the model intercepts. Predictions are based on separate GAMs fitted to the Beijing and Taiwan Mandarin datasets and are overlaid to facilitate comparison. Words that bear neutral tone in both varieties are shown in the upper panel. Words that are realized with neutral tone in Beijing Mandarin but not in Taiwan Mandarin are shown in the lower panel. In the lower panel, the pinyin transcription follows the Taiwan Mandarin standard.}
    \label{fig:word}
\end{figure}

Figure~\ref{fig:word} presents the predicted pitch contours for those that are attested in both the Beijing and Taiwan Mandarin datasets. The upper panel displays 27 words whose second syllable bears a neutral tone in both varieties. These words are organized by their morpho-syntactic structure, including reduplicated kinship terms such as 妈妈\ (\textit{ma1ma5}, `mom'), reduplicated verbs such as 看看\ (\textit{kan4kan5} `have a look'), plural suffix constructions (e.g., X+们 \textit{men5}, plural marker), nominal suffix constructions (e.g., X+子 \textit{zi5}, nominal suffix), structural particle constructions (e.g., X+的 \textit{de5}, structural particle), aspectual or other function-word constructions (e.g., 了 \textit{le5}, aspect marker), and lexicalized neutral-tone words such as 东西\ (\textit{dong1xi5}, `thing'). The lower panel presents 16 words the second syllable of which,  according to standard descriptions, is realized with a neutral tone in Beijing Mandarin but not in Taiwan Mandarin. 


Word-specific tone signatures are clearly visible when comparing words with the same tone pattern, such as 弟弟 (panel a, 4), 爸爸 (panel a, 6), 太太 (panel a, 27) and 记得 (panel a, 21).  Focusing on Beijing Mandarin (red curves), for 弟弟 we observe a clearly falling pitch contour, for 爸爸 a nearly level pitch contour, for 太太 the second syllable shows a clear rise, whereas for 记得, the pitch contour is inverse U-shaped. 

In the present dataset, there is one heterographic homophone:  \textit{ta1men5} (`they'), represented by 他们\ (`they, male or mixed gender') and 她们\ (`they, female'). In Beijing Mandarin (indicated by the red curve), 他们\ exhibits a relatively flat pitch contour, whereas 她们\ shows a slightly rising contour. 


Turning to Taiwan Mandarin (blue curves), some words have realizations that clearly diverge from the dictionary standards. For example, 早上\ (\textit{zao3shang5}, `morning', panel a, 25) is realized with a fall-rise-fall contour that can be interpreted as the dipping tone of 早 followed by the falling tone of 上 (as preposition or verb) rather than the neutral tone prescribed by the dictionary. The fall-rise-fall contour diverges markedly from the dipping contour of the T3-T5 pattern in Taiwan Mandarin (shown in Figure~\ref{fig:pattern}). 爷爷 (panel a, 5) shows a final rise that is much stronger than expected given the T2-T5 tone pattern (which is basically a mid level tone (see  Figure~\ref{fig:pattern}). 你们\ (\textit{ni3men5}, `you.PL', panel a, 9) displays a dipping contour at the end of the F0 trajectory, instead of a dipping contour spread out across two syllables.  Possibly, the final rise is due to rising tone of 们 (\textit{men2}, `door, opening').  For the one homophone pair in our dataset, in Taiwan Mandarin, 她们\ (\textit{ta1men5}, `they, female') is realized with a higher pitch than 他们\ (\textit{ta1men5}, `they, male or mixed gender'), consistent with patterns reported for the monosyllables 她\ (\textit{ta1}, `she') and 他\ (\textit{ta1}, `he') for this variety  \citep{jin2026new}. 


Words that are described as having a neutral tone in Beijing Mandarin and a lexical tone in Taiwan Mandarin also exhibit remarkable word-specific signatures.  Consider, for instance, 休息\ (\textit{xiu1xi2}, `rest', panel b, 15).  In Beijing Mandarin, this word is pronounced with an f0 contour that is similar to the pitch component of the T1-T5 tone pattern: an initial rise followed by a gentle fall.  For Taiwan Mandarin, instead of a T1-T2 tone pattern, we find a rise-fall pitch contour. For 告诉\ (\textit{gao4su4}, `tell', panel b, 2) and 厉害\ (\textit{li4hai4}, `serious/awesome', panel b, 3), the expected tone pattern is a rise-fall for Beijing Mandarin (see Figure~\ref{fig:pattern}, left panel), but for these words, the word specific pitch signatures superimpose a dipping pattern (告诉) or a general downward pitch trend (厉害). By contrast, in Taiwan Mandarin, there is no clear evidence for a falling pitch contour on the second syllable.  For 告诉, the overall pattern is that of a dipping tone that ends at the mid-to-low final pitch that is characteristic for this dialect.  However, for 厉害, the second syllable carries a weak dipping tone, rather than a falling tone. 

Words with the most markedly different tonal realizations are 爷爷 (panel a, 5), 她们 (panel a, 11), 样子 (panel a, 15), 早上 (panel a, 25) and 太太  (panel a, 27). More extreme differences are seen for words that have been reported to have floating tones in Beijing Mandarin and lexical tones in Taiwan Mandarin: 漂亮  (panel b, 6), 朋友 (panel b, 5), 厉害 (panel b, 3), 喜欢 (panel b, 13) and 休息 (panel b, 15).
%
%
%
%
%
%
But there are also words that exhibit remarkably similar contours across the two varieties. Examples include 妈妈\ (\textit{ma1ma5}, `mom', panel a, 1), 哥哥\ (\textit{ge1ge5}, `brother', panel a, 2), and 我们\ (\textit{wo3men5}, `we', panel a, 8). Even for 便宜\ (\textit{pian2yi5}, `cheap', panel b, 1), which dictionaries expect to differ across dialects, have fairly similar predicted tonal realizations. 



\begin{figure}[htbp]
    \centering
    \includegraphics[width=0.4\linewidth]{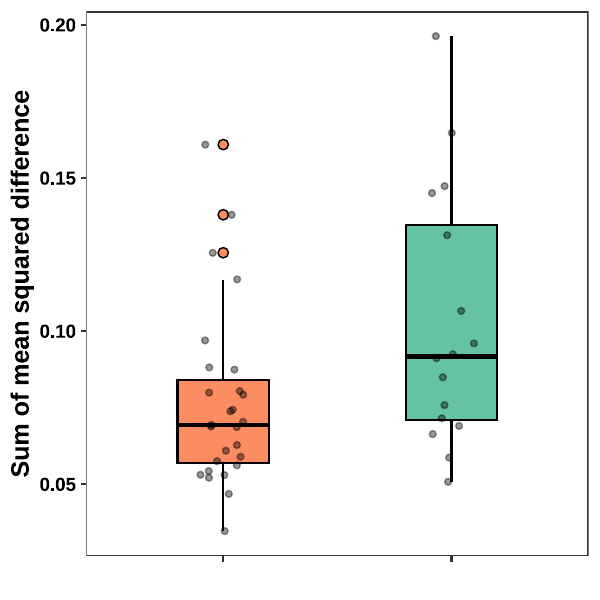}
    \caption{Distribution of mean squared difference between word type pairs. The orange box indicates words  bearing T5 in both Beijing and Taiwan Mandarin. The green box indicates words in which the second syllable is bearing T5 in Beijing Mandarin but not in Taiwan Mandarin.
    }
    \label{fig:boxplot}
\end{figure}

Overall, words bearing a neutral tone in both varieties exhibit more similar pitch contours than words that have a  neutral-toned only in Beijing Mandarin. This difference emerges clearly  from an inspection of the differences in the summed squared errors (SSD) between the predicted pitch contours of Beijing and Taiwan Mandarin for each word pair. Figure~\ref{fig:boxplot} shows the distribution of SSD values across the two groups. The orange box represents words that bear a neutral tone (T5) in both Beijing and Taiwan Mandarin, while the green box represents words whose second syllable bears a neutral tone in Beijing Mandarin but not in Taiwan Mandarin. The mean SSD for words bearing a neutral tone in both varieties was 0.07, whereas the mean SSD for words bearing a neutral tone only in Beijing Mandarin was 0.10 ($t = -2.2082$,  $df = 23.668$,  $p-value = 0.03717$).

In summary, with respect to the research question of whether \textbf{words have their own tonal signatures not only in Taiwan Mandarin but also in Beijing Mandarin}, the evidence unequivocally supports tonal signatures across both dialects.

\subsection{Meaning} \label{sec:modeling}

\noindent
In this section, we turn to the fourth and final research question, examining whether the word-specific tonal signatures observed in the preceding section are linked to the meanings of words in context. To do so, we calculated meaning vectors for the work tokens, and examined whether these meaning vectors are predictive for words' pitch contours.

%
%
%

\begin{figure}[htbp]
    \centering
    \includegraphics[width=\linewidth]{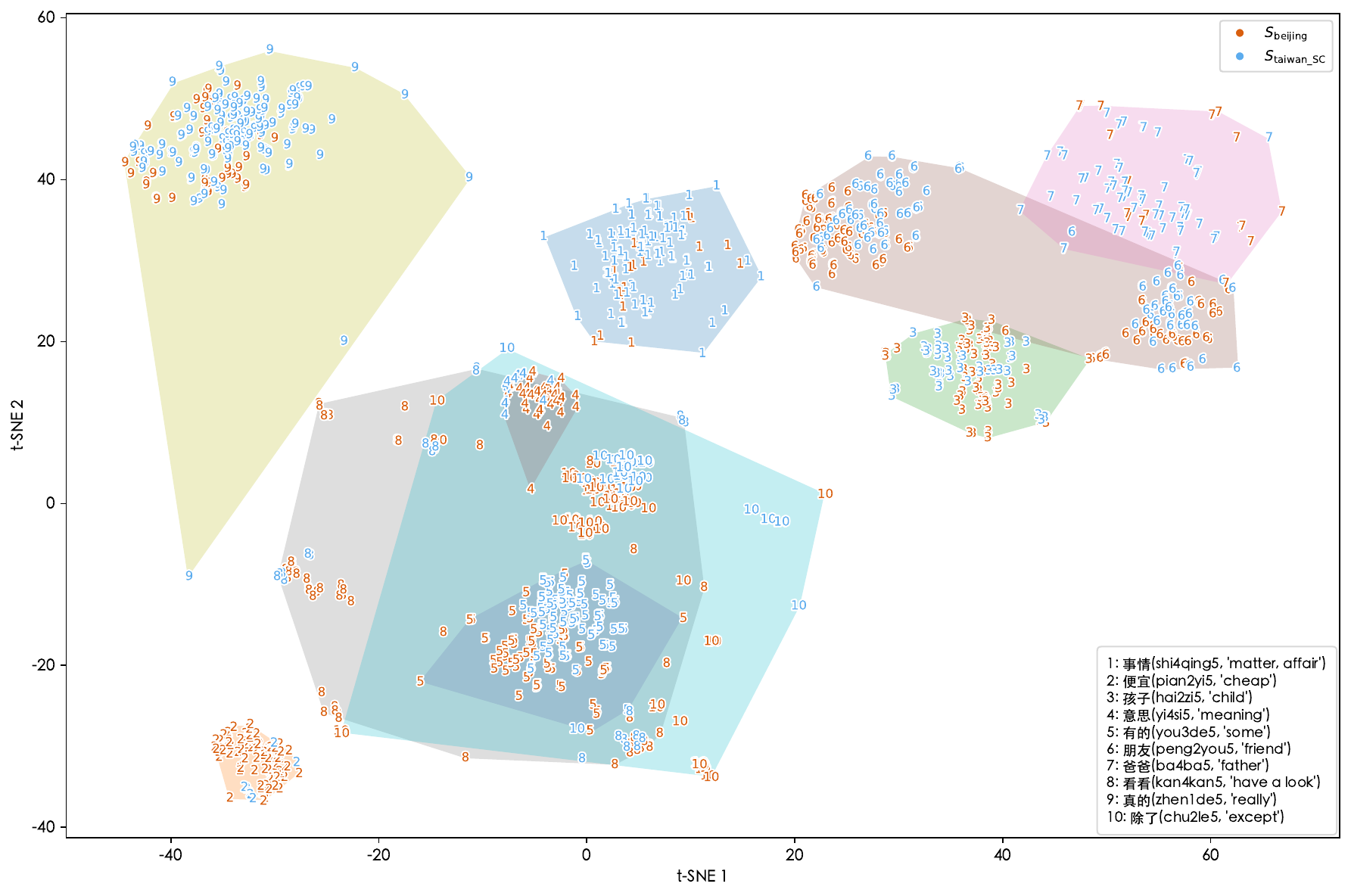}
    \caption{
    Contextualized embeddings $S_{\text{beijing}}$ and $S_{\text{taiwan\_SC}}$, obtained from a pretrained Chinese Qwen-2.5 model, are shown in a two-dimensional plane obtained with t-SNE. Convex hulls (polygons) highlight the clusters of the top-10 frequent word types, labeled by the numbers 1--10. Orange numbers represent tokens in $S_{\text{beijing}}$, and blue numbers represent tokens in $S_{\text{taiwan\_SC}}$.
    }
    \label{fig:embeddings}
\end{figure}

We operationalized words' meanings by means of contextualized embeddings, henceforth CEs. The embeddings in the current study were derived from Qwen-2.5, a family of large language models developed by Alibaba \citep{qwen2025qwen25technicalreport}. The embedding of a word token was computed for  the specific context in which that token occurred.  Thus, each word token was associated with an 896-dimensional vector. For the Beijing Mandarin dataset, which is annotated with simplified Chinese  characters, we generated a set of embeddings using simplified Chinese input. The embeddings were brought together as the row vector of a matrix  $S_{\text{beijing}}$. For the Taiwan Mandarin dataset, which is annotated in traditional Chinese characters, we generated two sets of embeddings, $S_{\text{taiwan\_TC}}$, based on the original traditional Chinese annotations, and $S_{\text{taiwan\_SC}}$, obtained by converting traditional characters into simplified Chinese before extracting embeddings. Each of these three matrices have dimensions $n \times 896$, where $n$ is the number of tokens in the dataset. $S_{\text{taiwan\_TC}}$ is used as the semantic representation of the Taiwan Mandarin dataset in the subsequent modeling. $S_{\text{taiwan\_SC}}$ is used only for the visualization in Figure~\ref{fig:embeddings}. Since $S_{\text{beijing}}$ and $S_{\text{taiwan\_SC}}$ reside in the same semantic space --- they are both calculated for simplified characters --- we can consider these CEs jointly. 

For tracing possible differences in meanings across the two dialects,  we reduced the 896-dimensional semantic space to two dimensions using t-SNE \citep{van2008visualizing}. Figure~\ref{fig:embeddings} shows the 10 most frequent word types shared by both datasets, labeled with the numbers 1 -- 10. Tokens from $S_{\text{beijing}}$ are indicated by orange numbers, and tokens from $S_{\text{taiwan\_SC}}$ are indicated by blue numbers. Convex hulls are used to highlight clusters corresponding to different word types. 

For most words, a clear single cluster is present that contains tokens from both the Beijing and Taiwan corpora, the only exception being 朋友\ (\textit{peng2you3}, `friend'), which forms two distinct clusters in the upper right quadrant of the t-SNE plane. The cluster on the left, close to 1, corresponds to abstract or relational uses, where 朋友\ denotes the existence or formation of social ties (e.g., making friends or having friends). The cluster on the right reflects comitative uses, in which 朋友\ appears as a co-participant in social activities (e.g., going out or engaging in activities with friends). For example, in the relational sense, 朋友\ occurs in expressions such as 能交上朋友\ (`to be able to make friends') or 因为上学的一些朋友\ (`friends made through school'), where the focus is on the establishment or existence of social relationships rather than shared actions. By contrast, in the comitative sense, 朋友\ appears in contexts describing joint activities, such as 周末会专门约朋友出去 (`[one] would specifically arrange to meet friends on weekends') or 跟朋友一起逛街、运动之类的 (`going shopping or exercising together with friends'), where the emphasis is on shared experiences and social interaction.

It is noteworthy that within several clusters, the tokens of the two dialects are not randomly distributed. For the clusters of 有的 (5) and 除了 (10), for instance, the tokens from the Beijing corpus have higher values on the vertical dimension compared to the tokens from the Taiwan corpus, suggesting that there are measurable differences in the senses of these words in the two dialects.     

%
%

Pitch contours were represented as fixed-length vectors in matrices $C_{\text{beijing}}$ and $C_{\text{taiwan}}$, each of size $n \times 100$. Each row corresponds to one token, and the 100 columns of a row vector represent the 100 time-normalized F0 measurements for that token. These token-specific pitch vectors were denoised using the best-fit GAMs described in Section~\ref{sec:analyses}, which take into account a wide range of factors that co-determine tonal realization. Following \citet{lu2026realization}, we extracted the smooth for normalized time, the word-specific smooths, and other smooth terms, and added these to obtain denoised word-specific pitch signatures.

For each variety, the dataset was randomly split into a training dataset (90\%) and a testing dataset (10\%), with the constraint that the testing dataset set did not contain tokens of types that it had not encountered during training.  A linear mapping was trained to project contextualized embeddings ($S$) onto the corresponding pitch contours ($C$): $S_{\text{beijing}}$ to $C_{\text{beijing}}$ for Beijing Mandarin, and $S_{\text{taiwan\_SC}}$ to $C_{\text{taiwan}}$ for Taiwan Mandarin. Model performance was evaluated using a nearest-neighbor classification approach. For the predicted contour of a token, its closest neighbor was identified using L2 distance. A prediction was considered as correct if the nearest neighbor matched the target word type; otherwise, it was considered as incorrect. The entire procedure was repeated over 30 random permutations, and mean accuracy was calculated.

The mean accuracy for Beijing Mandarin was 36.00\% for the training dataset and 23.97\% for the testing dataset (30-run global permutation baseline: 10.01\%). For Taiwan Mandarin, the mean accuracy across 30 permutations was 33.32\% for the training dataset and 20.34\% for the testing dataset, with a permutation baseline of 10.95\%. In both datasets, performance was well above the permutation baseline, indicating that tonal realizations of neutral tone can be predicted by their meaning in context with above-chance accuracy. 

\begin{figure}[htbp]
    \centering
    \begin{subfigure}[b]{\linewidth}
        \includegraphics[width=\linewidth]{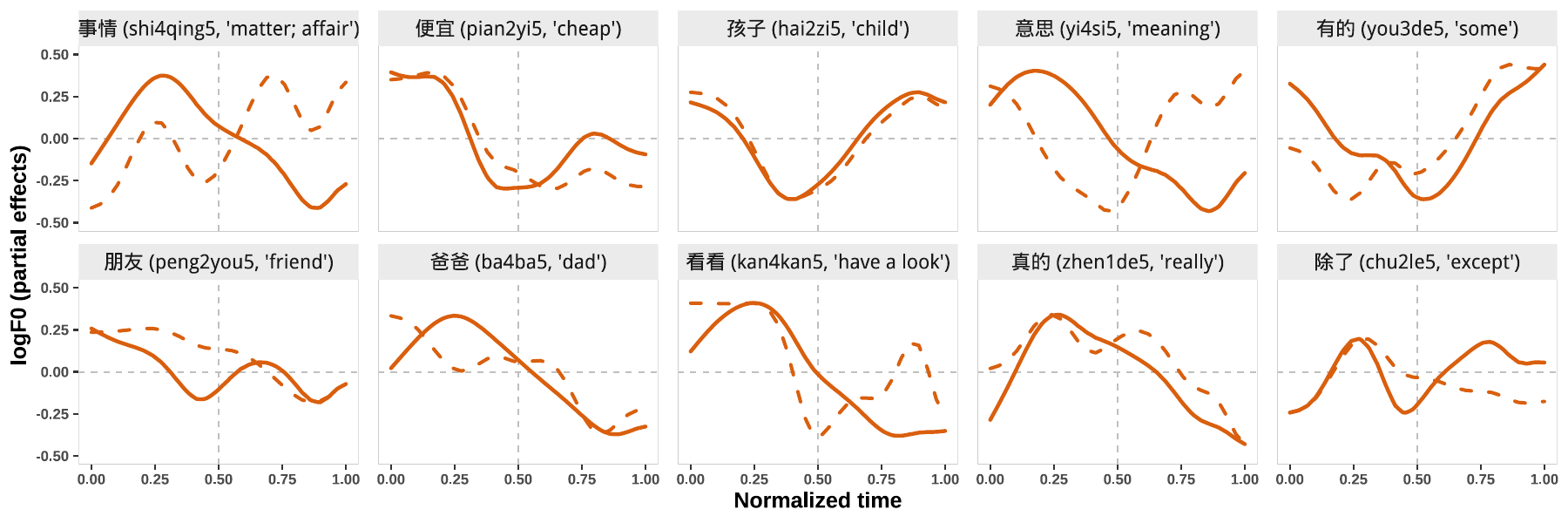}
        \caption{}
    \end{subfigure}
    \hfill
    \begin{subfigure}[b]{\linewidth}
        \includegraphics[width=\linewidth]{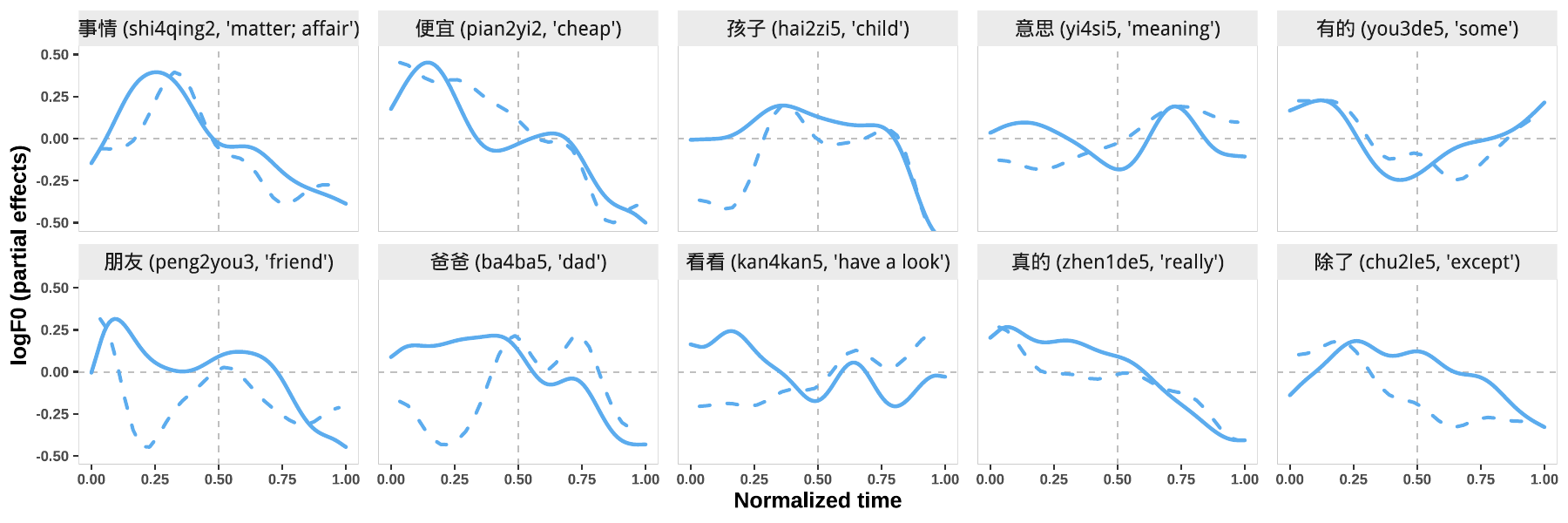}
        \caption{}
    \end{subfigure}
    \caption{
    Comparison of observed (dashed lines) and predicted pitch contours (solid lines) for the ten words shown in Figure~\ref{fig:embeddings}. The upper panel presents results for Beijing Mandarin, and the lower panel presents results for Taiwan Mandarin. Predicted contours were obtained by using the centroid of the word as input to the linear mapping. Observed contours are reproduced from the predicted pitch contours shown in Figure~\ref{fig:word} and rescaled to the same scale as the DLM predictions. 
    }
    \label{fig:chat_word}
\end{figure}

Figure~\ref{fig:chat_word} compares observed and predicted pitch contours for the ten words shown in Figure~\ref{fig:embeddings}. Predicted contours were obtained by giving the centroid embedding of each word as an input to the linear mapping. For both Beijing and Taiwan Mandarin, the DLM-predicted contours generally resemble the observed trajectories, although the degree varies across words. Some words exhibit similar contour shapes across the two varieties, such as 事情 and 有的, whereas others show more cross-variety differences in their observed realizations (e.g., \textit{孩子}). Overall, the results indicate that the embedding-based representations capture substantial information about the shapes of word-specific pitch contours.

\begin{figure}[htbp]
    \centering
    \begin{subfigure}[a]{\linewidth}
        \includegraphics[width=\linewidth]{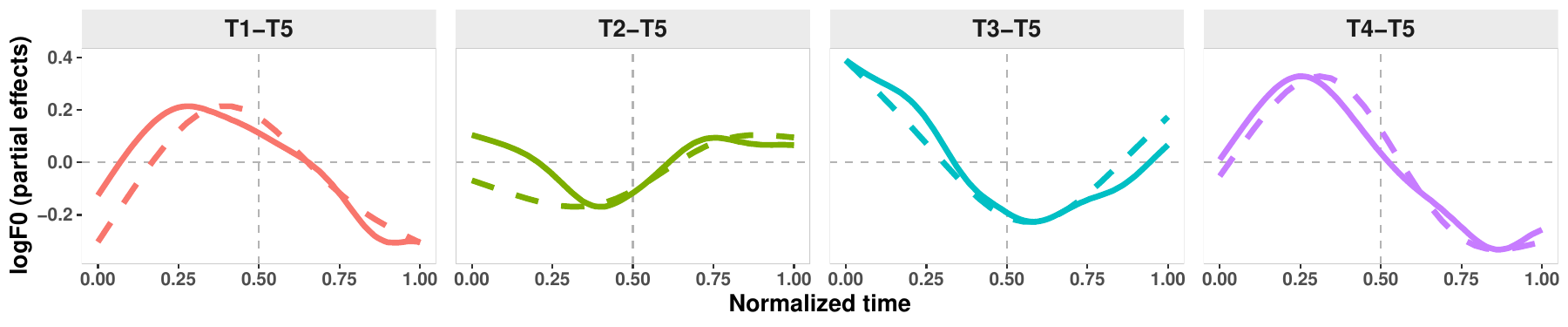}
        \caption{}
    \end{subfigure}
    \hfill
    \begin{subfigure}[b]{\linewidth}
        \includegraphics[width=\linewidth]{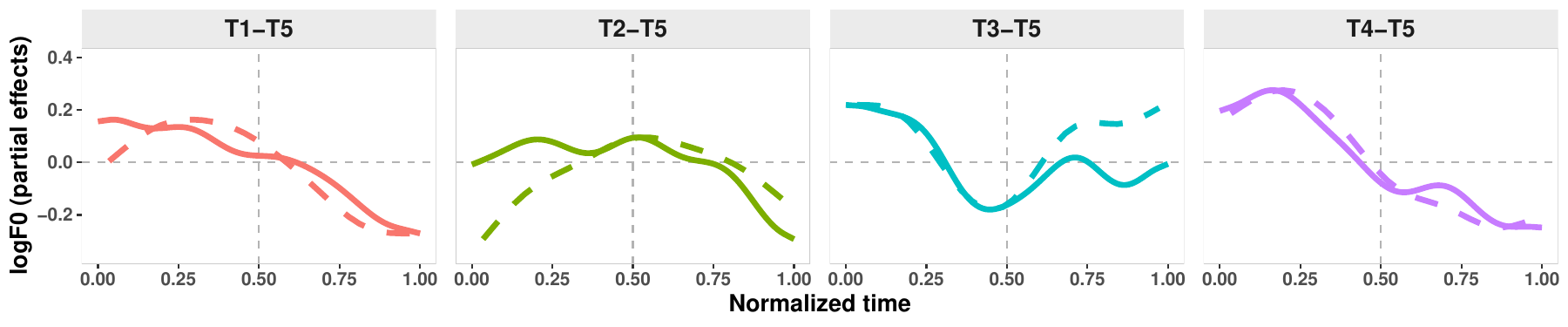}
        \caption{}
    \end{subfigure}
    \caption{
    CE-predicted pitch contours for tone patterns (solid lines) and GAM-estimated partial effects for tone pattern (dashed lines) for Beijing Mandarin (upper panel) and Taiwan Mandarin (lower panel). 
    For ease of comparison, the GAM-predicted contours are linearly rescaled to match the range of the CE-predicted contours. 
    }
    \label{fig:chat}
\end{figure}

In addition, we computed for each tone pattern the centroids of the CEs of the tokens belonging to that tone pattern. For each dialect separately, the resulting four centroid vectors, the `prototypical meanings' of the tone patterns, were given as input to the linear mapping, resulting in four predicted pitch contours.  Figure~\ref{fig:chat} compares these CE-predicted contours (solid lines) with the partial effect pitch curves obtained with the best-fit GAMs for tone pattern (dashed lines, cf. Figure~\ref{fig:pattern}). Two things are noteworthy. First, there is considerable similarity between the CE-predicted curves and the GAM-based partial effect curves. Given that the CEs are obtained with a large language model that is not tuned to the individual experiences of the speakers in the two corpora, perfect fits cannot be expected. Second, some of the differences between the tone-pattern contours of Beijing and Taiwan Mandarin may be due to subtle differences in the meanings-in-context of the word tokens in the two dialects, as exemplified in Figure~\ref{fig:embeddings}, especially in the case of the T1-T5 and T4-T5 tone patterns, for which the fits of the CE-predicted contours and the GAM-based observed contours are very similar within each dialect while being clearly different between dialects.

With respect to our fourth research question, which asks whether \textbf{contextualized embeddings for Beijing conversational speech are predictive for words' tonal signatures as previously observed for Taiwan Mandarin}, our results provide clear support for the possibility that words' pitch contours are in part co-determined by their meanings. Furthermore,  we have shown that the differences in the realization of the tone patterns across the two dialects may in part arise due to dialect-specific differences in words' contextualized meanings.

\section{General discussion}\label{sec:discussion}

\noindent
The neutral, or floating, tone of Mandarin Chinese is an enigmatic tone, that has been described as a reduced tone, a flexible tone that can be on a continuum from being a lexical tone to toneless, and as a tone that is highly context-dependent \citep[see, e.g.,][]{dong2025neutral}. In this study, we investigated the realization of two-syllable words with a neutral second tone, in corpora of conversational Mandarin recorded in Beijing and in Taiwan, focusing on the following four questions.
\begin{enumerate}
    \item \textbf{What is the nature of the neutral/floating tone?} For two-syllable word, the floating tone turns out to be characterized by a mid-to-low final tone target, which dovetails well with the findings of \citet{lee2003phonetic,huang2018phonological}. At the same time, each tone pattern has its own tonal characteristics. Therefore, the floating tone in two-syllable words is not different in principle from other lexical tones in second syllables \citep[cf.][]{lu2026realization}.  
    {\color{black}One possible interpretation is that the neutral tone is not a single lexical tone but rather a mixture of several distinct underlying tones. However, this possibility is not unique to the neutral tone and may also apply to other lexical tones. More importantly, when the neutral tone is treated as a single lexical category, as we do here, it exhibits the same kind of systematic variability as the other lexical tones, with an underlying unifying tonal pattern. This leads to the conclusion that the neutral (or floating) tone is a lexical tone, characterized by a mid-to-low final F0 target.}
    
    \item \textbf{Do tone patterns of words with the floating tone differ between Beijing and Taiwan Mandarin?}  A comparison of Beijing and Taiwan Mandarin revealed that words with the T1-T5 and T4-T5 tone patterns have an initial rise that is absent in Taiwan Mandarin.  In Taiwan Mandarin, words with the T2-T5 tone pattern are realized with a nearly flat tone contour, which contrasts with solid evidence for a rise in the T2-T5 words as realized in the Beijing corpus. Across the two dialects, the realization of the T3-T5 is most similar, and resembles a drawn-out dipping tone.  Furthermore, there is substantially more variability in the realization of words with a floating tone in Taiwan Mandarin, as compared to Beijing Mandarin. The high variability that characterizes Taiwan floating tones is consistent with the findings of, e.g., \citet{huang2018phonological}.    
    \item \textbf{Do words with a floating tone have word-specific tonal signatures?} In line with previous corpus-based studies of the realization of tones in conversational Mandarin \citep{Chuang:Bell:Tseng:Baayen:2026,lu2026realization,jin2026new}, solid evidence was obtained for word-specific pitch signatures, replicating \citet{lu2026realization} for Taiwan Mandarin, and extending the evidence to another variety, Beijing Mandarin. The variable importance of word-specific smooths was even greater for Beijing Mandarin as compared to Taiwan Mandarin.  Word-specific tone signatures can be much stronger than the pitch components tied to tone patterns, which helps explain why \citet{dong2025neutral} concluded that the neutral tone is a very flexible tone shaped by a variety of contextual factors. Importantly, the present study takes a wide range of contextual factors into account, including the substantial effect of the tones of adjacent syllables in the context.  Nevertheless, thanks to the ability of the generalized additive model to decompose a pitch contour into an additive set of partial contours that are tied to linguistic predictors, it is now possible to isolate in empirical pitch contours not only the components that are due to neighboring tones, position in the sentence, speaker, and probability in context, but also components linked to tone pattern and, importantly, word identity. 
    \item \textbf{Are word-specific pitch signatures predictable from their meanings in context?} As previous studies have provided evidence that word specific pitch signatures are linked to the meanings of words in context, following \citet{Chuang:Bell:Tseng:Baayen:2026,lu2026realization,jin2026new}, we examined whether words' contextualized embeddings are predictive for words' pitch signatures. A simple linear transformation from contextualized embeddings to pitch contours indeed predicts words' signatures with an accuracy that far exceeds a permutation baseline, in line with the predictions of the Discriminative Lexicon Model \citep{baayen_discriminative_2019,heitmeier2026discriminative}.  Furthermore, the prototypical meanings of the tone patterns, estimated by the centroids of their contextualized embeddings, predict pitch contour signatures that are remarkably similar to the pitch components isolated by the GAM for the tone patterns \citep[see also][]{Chuang:Bell:Tseng:Baayen:2026,lu2026realization}. Similarities are especially striking for the T1-T5 and T4-T5 tone patterns, and capture well the differences in the realization of these tone patterns in the two dialects.  This points to the possibility that differences in the realization of tone between the two dialects are in part due to differences in the senses of these words in context. 
\end{enumerate}

The actual tonal realizations in the corpus of conversational Taiwan Mandarin diverge considerably from the Taiwan dictionary norms. Words that are described as having a lexical tone on the second syllable (corresponding to a neutral tone in standard Mandarin and Beijing Mandarin) can have pitch contours in colloquial language use that bear no resemblance to the lexical tones found in the dictionary, while also diverging considerably from their tone-patterns as identified by the GAM.  Compared to the conversational Beijing Mandarin sampled by our corpus, there is more cross-dialect variation for words marked as having a neutral tone in this dialect, but marked as having a lexical tone in Taiwan, compared to words that are described as having a neutral tone in both dialects.

Two directions for future research seem especially promising. First, the pitch contour is not the only acoustic correlate of the neutral tone. Neutral tones, at least in laboratory speech, tend to have shorter duration. Future work could therefore examine whether differences in meaning can be reflected in duration. Evidence for the effect of meaning (using embeddings) on spoken word duration has been reported for English \citep{Gahl:Baayen:2024}.  Second, previous studies have investigated single-syllable words \citep{jin2026new} and bisyllabic words \citep{Chuang:Bell:Tseng:Baayen:2026,lu2026realization}, but it is currently unclear whether word-specific pitch signatures are also present for three-syllable words. Previous research suggests that the pitch realization of trisyllabic sequences differs from that of disyllabic forms \citep{xu2024cross}. Specifically tri-syllabic sequences with two adjacent neutral tones will likely be both very interesting, but also highly challenging due to data sparsity in current corpora of spontaneous conversational speech.


To conclude, the present investigation shows that in colloquial conversational Mandarin as spoken in Beijing and Taiwan, the neutral or floating tone as found in two-syllable words is highly similar to the lexical tones in many ways. The neutral tone has its own pitch target, just as do the other tones. Words with the neutral tone have tone patterns, exactly as observed for the lexical tone.  Words with the neutral tone have their own pitch signatures, which are to some extent predictable from their contextualized embeddings, mirroring what has been found for the lexical tones.  The only way in which neutral tones differ from lexical tones is that neutral tones in di-syllabic words are restricted to the second syllable. From a meta-theoretical perspective, there is one further difference: the neutral tone is the only tone the variability of which has been extensively commented on, even though similar variability is widespread among the lexical tones.

\section*{Funding}
This work was supported by the European Research Council under Grant SUBLIMINAL (\#101054902) awarded to R. Harald Baayen.

\section*{Declaration}
The authors declare no conflicts of interest.

\section*{Data availability}
The data that support the findings of this study are available in an OSF repository at \url{https://osf.io/cre4z/overview?view_only=7416dd299535463781e51389d47a1fd8}.

\newpage
\section*{Appendix} \label{sec:appendix}
\bigbreak

\renewcommand{\thetable}{A.\arabic{table}}
\setcounter{table}{0} 

\renewcommand{\thefigure}{A.\arabic{figure}}
\setcounter{figure}{0} 

\begin{table}[h]
\centering
\caption{Model summary of best-fit GAM fitted to the dataset of Beijing Mandarin.} 
\begin{tabular}{lrrrr}
\hline
A. parametric coefficients & Estimate & Std. Error & t-value & p-value \\ 
  (Intercept) & 5.1939 & 0.0420 & 123.7213 & $<$ 0.0001 \\ 
  tone\_pattern25 & -0.0956 & 0.0170 & -5.6307 & $<$ 0.0001 \\ 
  tone\_pattern35 & -0.1537 & 0.0185 & -8.3203 & $<$ 0.0001 \\ 
  tone\_pattern45 & -0.0352 & 0.0158 & -2.2337 & 0.0255 \\ 
   \hline
B. smooth terms & edf & Ref.df & F-value & p-value \\ 
  s(normalized\_t):tone\_pattern15 & 3.7130 & 3.7931 & 28.1303 & $<$ 0.0001 \\ 
  s(normalized\_t):tone\_pattern25 & 3.7615 & 3.8193 & 15.4879 & $<$ 0.0001 \\ 
  s(normalized\_t):tone\_pattern35 & 3.7193 & 3.7915 & 31.2173 & $<$ 0.0001 \\ 
  s(normalized\_t):tone\_pattern45 & 3.9244 & 3.9454 & 67.8569 & $<$ 0.0001 \\ 
  s(normalized\_t,word) & 713.3847 & 905.0000 & 13.5714 & $<$ 0.0001 \\ 
  s(normalized\_t,speaker) & 350.2451 & 449.0000 & 65.6793 & $<$ 0.0001 \\ 
  s(normalized\_t,tonal\_context) & 278.8765 & 323.0000 & 17.9296 & $<$ 0.0001 \\ 
  s(logdur) & 6.3463 & 6.8575 & 21.9195 & $<$ 0.0001 \\ 
  ti(normalized\_t,logdur) & 14.8811 & 15.7724 & 34.5459 & $<$ 0.0001 \\ 
  s(norm\_utt\_pos) & 3.8331 & 3.9744 & 416.5007 & $<$ 0.0001 \\ 
  ti(normalized\_t,norm\_utt\_pos) & 8.2397 & 8.8038 & 9.0758 & $<$ 0.0001 \\ 
  s(bg\_prob\_prev) & 3.9048 & 3.9856 & 30.5617 & $<$ 0.0001 \\ 
  ti(normalized\_t,bg\_prob\_prev) & 12.4668 & 14.3497 & 7.7244 & $<$ 0.0001 \\ 
  s(bg\_prob\_fol) & 3.8222 & 3.9669 & 23.9617 & $<$ 0.0001 \\ 
  ti(normalized\_t,bg\_prob\_fol) & 8.8918 & 10.6017 & 24.9885 & $<$ 0.0001 \\ 
  \hline
\end{tabular}
\label{tab.gam}
\end{table}

\clearpage

\begin{table}[h]
\centering
\caption{Model summary of best-fit GAM fitted to the dataset of Taiwan Mandarin.} 
\begin{tabular}{lrrrr}
\hline
A. parametric coefficients & Estimate & Std. Error & t-value & p-value \\ 
  (Intercept) & 5.1331 & 0.0457 & 112.2476 & $<$ 0.0001 \\ 
  tone\_pattern25 & -0.0915 & 0.0314 & -2.9175 & 0.0035 \\ 
  tone\_pattern35 & -0.1509 & 0.0327 & -4.6189 & $<$ 0.0001 \\ 
  tone\_pattern45 & -0.0305 & 0.0290 & -1.0517 & 0.2929 \\ 
  tone\_patternOthers & -0.0293 & 0.0279 & -1.0513 & 0.2931 \\ 
   \hline
B. smooth terms & edf & Ref.df & F-value & p-value \\ 
  s(normalized\_t):tone\_pattern15 & 3.1992 & 3.4331 & 1.5585 & 0.1887 \\ 
  s(normalized\_t):tone\_pattern25 & 3.1806 & 3.4278 & 1.4273 & 0.2195 \\ 
  s(normalized\_t):tone\_pattern35 & 4.5887 & 4.9074 & 2.4333 & 0.0391 \\ 
  s(normalized\_t):tone\_pattern45 & 4.8268 & 5.1607 & 4.3282 & 0.0006 \\ 
  s(normalized\_t):tone\_patternOthers & 6.6323 & 6.7192 & 12.3851 & $<$ 0.0001 \\ 
  s(normalized\_t,word) & 498.8511 & 718.0000 & 12.4295 & $<$ 0.0001 \\ 
  s(normalized\_t,speaker) & 399.3609 & 494.0000 & 415.2898 & $<$ 0.0001 \\ 
  s(normalized\_t,tonal\_context) & 277.1922 & 323.0000 & 24.5113 & $<$ 0.0001 \\ 
  s(logdur) & 5.7031 & 5.9535 & 179.3700 & $<$ 0.0001 \\ 
  ti(normalized\_t,logdur) & 8.1978 & 8.7727 & 30.1564 & $<$ 0.0001 \\ 
  s(norm\_utt\_pos) & 5.9299 & 5.9965 & 572.4320 & $<$ 0.0001 \\ 
  ti(normalized\_t,norm\_utt\_pos) & 8.2106 & 8.7625 & 13.8231 & $<$ 0.0001 \\ 
  s(bg\_prob\_prev) & 2.9323 & 2.9929 & 248.7163 & $<$ 0.0001 \\ 
  ti(normalized\_t,bg\_prob\_prev) & 14.6629 & 15.5636 & 17.0239 & $<$ 0.0001 \\ 
  s(bg\_prob\_fol) & 2.9944 & 2.9996 & 199.8051 & $<$ 0.0001 \\ 
  ti(normalized\_t,bg\_prob\_fol) & 13.3945 & 14.9893 & 10.0016 & $<$ 0.0001 \\ 
  \hline
\end{tabular}
\label{tab.gam}
\end{table}

\clearpage
\printbibliography

\end{document}